\documentclass{article} % For LaTeX2e
\usepackage{iclr2021_conference,times}

\usepackage{amsmath,amsfonts,bm}

\def\Figref#1{Figure~\ref{#1}}

\def\Secref#1{Section~\ref{#1}}
\def\eqref#1{equation~\ref{#1}}
\def\Eqref#1{Equation~\ref{#1}}
\def\1{\bm{1}}

\DeclareMathAlphabet{\mathsfit}{\encodingdefault}{\sfdefault}{m}{sl}
\SetMathAlphabet{\mathsfit}{bold}{\encodingdefault}{\sfdefault}{bx}{n}

\usepackage{hyperref}
\usepackage{url}
\usepackage{array}

\usepackage{soul}
\usepackage{xcolor}

\usepackage{inconsolata}
\definecolor{light-gray}{gray}{0.95}
\newcommand{\code}[1]{\colorbox{light-gray}{\texttt{#1}}}

\newcommand{\sphere}{\ensuremath{{\mathbb{S}^2}}}
\newcommand{\sothree}{\ensuremath{{\mathrm{SO}(3)}}}
\newcommand{\fset}{\ensuremath{{\boldsymbol{\mathcal{F}}_L}}}
\newcommand{\fsetk}{\ensuremath{{\boldsymbol{\mathcal{F}}^K_L}}}
\newcommand{\ntp}{\ensuremath{{\mathcal{N}_{\otimes}}}}
\newcommand{\plset}{\ensuremath{{\mathbb{P}^{\ell}_L}}}

\newcommand{\rot}{\ensuremath{{\mathcal{R}_{\rho}}}}
\newcommand{\aleg}[3]{\ensuremath{P_{#1}^{#2}({#3})}}

\usepackage{mathrsfs}
\usepackage{mathabx}
\usepackage{graphicx}
\usepackage{subcaption}
\usepackage[export]{adjustbox}

\title{Efficient Generalized Spherical CNNs}

\author{Oliver J.~Cobb, Christopher G.~R.~Wallis,
Augustine N.~Mavor-Parker, \\
\textbf{Augustin Marignier, Matthew A.~Price, Mayeul d’Avezac
\& Jason D.~McEwen}\thanks{Corresponding author: \texttt{jason.mcewen@kagenova.com}} \\
Kagenova Limited, Guildford GU5 9LD, UK%\\
}

\iclrfinalcopy % Uncomment for camera-ready version, but NOT for submission.
\ificlrfinal
    \newcommand{\blinded}[1]{#1}
\else
    \newcommand{\blinded}[1]{}
\fi
\begin{document}

\maketitle

\begin{abstract}
    Many problems across computer vision and the natural sciences require the analysis of spherical data, for which representations may be learned efficiently by encoding equivariance to rotational symmetries.  We present a generalized spherical CNN framework that encompasses various existing approaches and allows them to be leveraged alongside each other.  The only existing non-linear spherical CNN layer that is strictly equivariant has complexity $\mathcal{O}(C^2L^5)$, where $C$ is a measure of representational capacity and $L$ the spherical harmonic bandlimit.  Such a high computational cost often prohibits the use of strictly equivariant spherical CNNs.  We develop two new strictly equivariant layers with reduced complexity $\mathcal{O}(CL^4)$ and $\mathcal{O}(CL^3 \log L)$, making larger, more expressive models computationally feasible.  Moreover, we adopt efficient sampling theory to achieve further computational savings. We show that these developments allow the construction of more expressive hybrid models that achieve state-of-the-art accuracy and parameter efficiency on spherical benchmark problems.
\end{abstract}

\section{Introduction}

Many fields involve data that live inherently on spherical manifolds, e.g.\ 360${^\circ}$ photo and video content in virtual reality and computer vision, the cosmic microwave background radiation from the Big Bang in cosmology, topographic and gravitational maps in planetary sciences, and molecular shape orientations in molecular chemistry, to name just a few.  Convolutional neural networks (CNNs) have been tremendously effective for data defined on Euclidean domains, such as the 1D line, 2D plane, or nD volumes, thanks in part to their translation invariance properties. However, these techniques are not effective for data defined on spherical manifolds, which have a very different geometric structure to Euclidean spaces (see Appendix~\ref{appendix:sphere}). To transfer the remarkable success of deep learning to data defined on spherical domains, deep learning techniques defined inherently on the sphere are required.  Recently, a number of spherical CNN constructions have been proposed.

Existing CNN constructions on the sphere fall broadly into three categories: fully real (i.e.\ pixel) space approaches \citep[e.g.][]{NIPS2017_6935,jiang2019spherical,perraudin2019deepsphere,cohen2019gauge}; combined real and harmonic space approaches \citep{cohen2018spherical,esteves2018learning,esteves2020spin}; and fully harmonic space approaches \citep{kondor2018clebsch}.
Real space approaches can often be computed efficiently but they necessarily provide an approximate representation of spherical signals and the connection to the underlying continuous symmetries of the sphere is lost.  Consequently, such approaches cannot fully capture rotational equivariance.
Other constructions take a combined real and harmonic space approach \citep{cohen2018spherical,esteves2018learning,esteves2020spin}, where sampling theorems \citep{driscoll:1994,kostelec:2008} are exploited to connect with underlying continuous signal representations to capture the continuous symmetries of the sphere.  However, in these approaches non-linear activation functions are computed pointwise in real space, which induces aliasing errors that break strict rotational equivariance.
Fully harmonic space spherical CNNs have been constructed by \citet{kondor2018clebsch}.  A continual connection with underlying continuous signal representations is captured by using harmonic signal representations throughout.  Consequently, this is the only approach exhibiting strict rotational equivariance.  However, strict equivariance comes at great computational cost, which can often prohibit usage.

In this article we present a generalized framework for CNNs on the sphere (and rotation group), which encompasses and builds on the influential approaches of \citet{cohen2018spherical}, \citet{esteves2018learning} and \citet{kondor2018clebsch} and allows them to be leveraged alongside each other. We adopt a harmonic signal representation in order to retain the connection with underlying continuous representations and thus capture all symmetries and geometric properties of the sphere.  We construct new fully harmonic (non-linear) spherical layers that are strictly rotationally equivariant, are parameter-efficient, and dramatically reduce computational cost compared to similar approaches.  This is achieved by a channel-wise structure, constrained generalized convolutions, and an optimized degree mixing set determined by a minimum spanning tree.  Furthermore, we adopt efficient sampling theorems on the sphere \citep{mcewen2011novel} and rotation group \citep{mcewen2015novel} to improve efficiency compared to the sampling theorems used in existing approaches \citep{driscoll:1994,kostelec:2008}.  We demonstrate state-of-the-art performance on all spherical benchmark problems considered, both in terms of accuracy and parameter efficiency.

%%% Local Variables:
%%% mode: latex
%%% TeX-master: "iclr2021"
%%% End:

\section{Generalized Spherical CNNs}\label{sec:gs2cnn}

We first overview the theoretical underpinnings of the spherical CNN frameworks introduced by \citet{cohen2018spherical}, \citet{esteves2018learning}, and \cite{kondor2018clebsch}, which make a connection to underlying continuous signals through harmonic representations. For more in-depth treatments of the underlying harmonic analysis we recommend \cite{esteves2020theoretical}, \cite{kennedy2013hilbert} and \cite{gallier2019aspects}. We then present a generalized spherical layer in which these and other existing frameworks are encompassed, allowing existing frameworks to be easily integrated and leveraged alongside each other in hybrid networks.

Throughout the following we consider a network composed of $S$ rotationally equivariant layers $\mathcal{A}^{(1)},....,\mathcal{A}^{(S)}$, where the $i$-th layer $\mathcal{A}^{(i)}$ maps an input activation $f^{(i-1)} \in \boldsymbol{\mathcal{H}}^{(i-1)}$ onto an output activation \mbox{$f^{(i)} \in \boldsymbol{\mathcal{H}}^{(i)}$}.
We focus on the case where the network input space $\boldsymbol{\mathcal{H}}^{(0)}$ consists of spherical signals (but note that input signals on the rotation group may also be considered).

\subsection{Signals on the Sphere and Rotation Group}\label{sec:gs2cnn:signals}

Let $L^2(\Omega)$ denote the space of square-integrable functions over domain $\Omega$. A signal $f \in L^2(\Omega)$ on the sphere ($\Omega=\sphere$) or rotation group ($\Omega=\sothree$) can be rotated by $\rho \in \sothree$ by defining the action of rotation on signals by $\rot f(\omega)=f(\rho^{-1}\omega)$ for $\omega \in \Omega$. An operator $\mathcal{A}: L^2(\Omega_1) \rightarrow L^2(\Omega_2)$, where $\Omega_1, \Omega_2 \in \{\sphere, \sothree\}$, is then equivariant to rotations if $\rot(\mathcal{A}(f))=\mathcal{A}(\rot f)$ for all $f\in L^2(\Omega_1)$ and $\rho \in \sothree$, i.e.\ rotating the function before application of the operator is equivalent to application of the operator first, followed by a rotation.

A spherical signal $f \in L^2(\sphere)$ admits a harmonic representation  $(\hat{f}^0,\hat{f}^1,...,)$ where $\hat{f}^{\ell} \in \mathbb{C}^{2\ell+1}$ are the harmonic coefficients given by the inner product $\langle f, Y^{\ell}_m \rangle$, where $Y^{\ell}_m$ are the spherical harmonic functions of degree $\ell$ and order $| m | \leq \ell$.
Likewise a signal $f \in L^2(\sothree)$ on the rotation group admits a harmonic representation $(\hat{f}^0,\hat{f}^1,...)$ where $\hat{f}^{\ell} \in \mathbb{C}^{(2\ell+1) \times (2\ell+1)}$ are the harmonic coefficients with $(m,n)$-th entry $\langle f,D^{\ell}_{mn} \rangle$ for integers $|m|,|n| \leq \ell$, where \mbox{$D^{\ell}: \sothree \rightarrow \mathbb{C}^{(2\ell+1)\times(2\ell+1)}$} is the unique $2\ell+1$ dimensional irreducible group representation %\footnote{To avoid any confusion we are explicit when using `representation' to refer to representations of groups.}
of \sothree\ on $\mathbb{C}^{(2\ell+1)}$.
The rotation $f \mapsto \rot f$ of a signal $f \in L^2(\Omega)$ can be described in harmonic space by $\hat{f}^{\ell} \mapsto D^{\ell}(\rho)\hat{f}^{\ell}$.

A signal on the sphere or rotation group is said to be bandlimited at $L$ if, respectively, $\langle f, Y^{\ell}_m \rangle=0$ or $\langle f, D^{\ell}_{mn} \rangle=0$ for $\ell \geq L$. Furthermore, a signal on the rotation group is said to be azimuthally bandlimited at $N$ if, additionally, $\langle f, D^{\ell}_{mn} \rangle=0$ for $|n|\geq N$. Bandlimited signals therefore admit finite harmonic representations $(\hat{f}^0,...,\hat{f}^{L-1})$. In practice real-world signals can be accurately represented by suitably bandlimited signals; henceforth, we assume signals are bandlimited.

\subsection{Convolution on the Sphere and Rotation Group}\label{sec:gs2cnn:conv}

A standard definition of convolution between two signals $f, \psi \in L^2(\Omega)$ on either the sphere ($\Omega=\sphere$) or rotation group ($\Omega =\sothree$) is given by
\begin{equation}\label{eqn:g-conv}
  (f \star \psi)(\rho) = \langle f, \rot\psi\rangle = \int_{\Omega} \text{d}\mu(\omega) f(\omega) \, \psi^*(\rho^{-1}\omega),
\end{equation}
where $d\mu(\omega)$ denotes the Haar measure on $\Omega$ and ${\cdot}^*$ complex conjugation \citep[e.g.][]{wandelt:2001,mcewen:2006:fcswt,mcewen:2013:waveletsxv,mcewen:s2let_spin,mcewen:s2let_localisation,cohen2018spherical,esteves2018learning}.  In particular, the convolution satisfies
\begin{equation}
((\rot f) \star \psi)(\rho') = \langle \mathcal{R}_{\rho} f, \mathcal{R}_{\rho'}\psi\rangle = \langle f, \mathcal{R}_{\rho^{-1}\rho'}\psi\rangle = (\mathcal{R}_{\rho} (f \star \psi))(\rho')
\end{equation}
and is therefore a rotationally equivariant linear operation, which we shall denote by $\mathcal{L}^{(\psi)}$.

The convolution of bandlimited signals can be computed exactly and efficiently in harmonic space as
\begin{equation}\label{eqn:conv}
  \widehat{(f \star \psi)}{}^{\ell} = \hat{f}^{\ell}  \hat{\psi}^{\ell*}, \hspace{50pt} \ell=0,...,L-1,
\end{equation}
which for each degree $\ell$ is a vector outer product for signals on the sphere and a matrix product for signals on the rotation group (see Appendix~\ref{appendix:convolutions} for further details). Convolving in this manner results in signals on the rotation group (for inputs on both the sphere and rotation group).  However, in the spherical case, if the filter is invariant to azimuthal rotations the resultant convolved signal may be interpreted as a signal on the sphere (see Appendix~\ref{appendix:convolutions}).

\subsection{Generalized Signal Representations}

The harmonic representations and convolutions described above have proven useful for describing rotationally equivariant linear operators $\mathcal{L}^{(\psi)}$. \cite{cohen2018spherical} and \cite{esteves2018learning} define spherical CNNs that sequentially apply this operator, with intermediary representations taking the form of signals on $\sothree$ and $\sphere$ respectively. Alternatively, for intermediary representations we now consider the more general space of signals introduced by \cite{kondor2018clebsch}, to which the aforementioned notions of rotation and convolution naturally extend.

In describing the generalization we first note from \Secref{sec:gs2cnn:signals} that all bandlimited signals on the sphere and rotation group can be represented as a set of variable length vectors of the form
\begin{equation}
  f=\{\hat{f}_t^{\ell} \in \mathbb{C}^{2\ell+1}: \ell=0,..,L-1;\; t=1,...,\tau_f^{\ell}\},
\end{equation}
where $\tau_f^{\ell}=1$ for signals on the sphere and \mbox{$\tau_f^{\ell}=\min(2\ell+1,2N-1)$} for signals on the rotation group. The generalization is to let \fset\ be the space of all such sets of variable length vectors, with $\tau_f$ unrestricted. This more general space contains the spaces of bandlimited signals on the sphere and rotation group as strict subspaces. For a generalized signal $f \in \fset$ we adopt the terminology of \citet{kondor2018clebsch} by referring to $\hat{f}^{\ell}_t$ as the $t$-th fragment of degree $\ell$ and to $\tau_f=(\tau_f^0, ...,\tau_f^{L-1})$, specifying the number of fragments for each degree, as the type of $f$.
The action of rotations upon \fset\ can be naturally extended from their action upon $L^2(\sphere)$ and $L^2(\sothree)$. For $f \in \fset$ we define the rotation operator $f \mapsto \rot f$ by $\hat{f}^{\ell}_t \mapsto D^{\ell}(\rho)\hat{f}^{\ell}_t$,
allowing us to extend the usual notion of equivariance to operators \mbox{$\mathcal{A}: \fset \rightarrow \fset$}.

\subsection{Generalized Convolutions}\label{sec:gs2cnn:gconv}

The convolution described by \Eqref{eqn:g-conv} provides a learnable linear operator $\mathcal{L}^{(\psi)}$ that satisfies the desired property of equivariance. Nevertheless, given the generalized interpretation of signals on \sphere\ and \sothree\ as signals in \fset, the notion of convolution can also be generalized \citep{kondor2018clebsch}.

In order to linearly and equivariantly transform a signal $f \in \fset$ of type $\tau_f$ into a new signal $f \ast \psi \in \fset$ of any desired type $\tau_{(f \ast \psi)}$, we may specify a filter $\psi = \{\hat{\psi}^{\ell} \in \mathbb{C}^{\tau^{\ell}_f\times \tau^{\ell}_{(f\ast\psi)}}: \ell=0,...,L-1\}$, which in general is not an element of \fset, and define a transformation $f \mapsto f\ast\psi$ by
\begin{equation}\label{eqn:linear-step}
  \widehat{(f \ast \psi)}^{\ell}_t = \sum_{t'=1}^{\tau^{\ell}_f} \hat{f}^{\ell}_{t'} \, \hat{\psi}^{\ell}_{t,t'}, \hspace{50pt} \ell = 0,..., L-1;\;\; t = 1,...,\tau^{\ell}_{(f\ast\psi)}.
\end{equation}
The degree-$\ell$ fragments of the transformed signal $(f\ast\psi)$ are simply linear combinations of the degree-$\ell$ fragments of $f$, with no mixing between degrees. Equation~\ref{eqn:conv} shows that this is precisely the form taken by convolution on the sphere and rotation group. In fact \cite{kondor2018generalization} show that all equivariant linear operations take this general form; the standard notion of convolution is just a special case. One benefit to the generalized notion is that the filter $\psi$ is not forced to occupy the same domain as the signal $f$, thus allowing control over the type $\tau_{(f \ast \psi)}$ of the transformed signal. We use $\mathcal{L}^{(\psi)}_\text{G}$ to denote this generalized convolutional operator.

\subsection{Non-linear Activation Operators}

For \fset\ to be a useful representational space, it must be possible to not only linearly but also non-linearly transform its elements in an equivariant manner.  However, equivariance and non-linearity is not enough. Equivariant linear operators cannot mix information corresponding to different degrees. Therefore it is of crucial importance that degree mixing is achieved by the non-linear operator.

\subsubsection{Pointwise Activations}\label{sec:gs2cnn:pw}

When the type $\tau_f$ of $f \in \fset$ permits an interpretation as a signal on \sphere\ or \sothree\ we may perform an inverse harmonic transform to map the function onto a sample-based representation \citep[e.g.][]{driscoll:1994,mcewen2011novel,kostelec:2008,mcewen2015novel}. A non-linear function $\sigma: \mathbb{C} \rightarrow \mathbb{C}$ may then be applied pointwise, i.e.\ separately to each sample, before performing a harmonic transform to return to a representation in \fset. We denote the corresponding non-linear operator as $\mathcal{N}_{\sigma}(f)=\mathcal{F}(\sigma(\mathcal{F}^{-1}(f)))$, where $\mathcal{F}$ represents the harmonic (i.e.\ Fourier) transform on \sphere\ or \sothree.  The computational cost of the non-linear operator is dominated by the harmonic transforms. While costly, fast algorithms can be leveraged (see Appendix~\ref{appendix:sphere}).  While inverse and forward harmonic transforms on \sphere\ or \sothree\ that are based on a sampling theory maintain perfect equivariance for bandlimited signals, the pointwise application of $\sigma$ (most commonly ReLU) is only equivariant in the continuous limit $L \rightarrow \infty$. For any finite bandlimit $L$, aliasing effects are introduced such that equivariance becomes approximate only, as shown by the following experiments.

We consider 100 random rotations $\rho \in \sothree$, for each of 100 random signal-filter pairs $(f,\psi)$, and compute the mean equivariance error $d(\mathcal{A}(\rot f), \rot (\mathcal{A} f))$ for operator $\mathcal{A}$, where $d(f,g)={\|f-g\|}/{\|f\|}$ is the relative distance between signals.  For convolutions the equivariance error is \mbox{$4.4 \times 10^{-7}$} for signals on \sphere\ and $5.3 \times 10^{-7}$ for signals on \sothree\ (achieving floating point precision). By comparison the equivariance error for a pointwise ReLU is $0.34$ for signals on \sphere\ and $0.37$ for signals on \sothree.  Only approximate equivariance is achieved for the ReLU since the non-linear operation spreads information to higher degrees that are not captured at the original bandlimit, resulting in aliasing.  To demonstrate this point we reduce aliasing error by oversampling the real-space signal.  When oversampling by $2\times$ or $8\times$ for signals on \sothree\ the equivariance error of the ReLU is reduced to $0.10$ and $0.01$, respectively.  See Appendix~\ref{appendix:equivariance} for further experimental details.

Despite the high cost of repeated harmonic transforms and imperfect equivariance, this is nevertheless the approach adopted by \citet{cohen2018spherical}, \citet{esteves2018learning} and others, who find empirically that such models maintain a reasonable degree of equivariance.

\subsubsection{Tensor-Product Activations}\label{sec:gs2cnn:tp}

In order to define a strictly equivariant non-linear operation that can be applied to a signal $f \in \fset$ of any type $\tau_f$ the decomposability of tensor products between group representations may be leveraged, as first considered by \cite{thomas2018tensor} in the context of neural networks.

Given two group representations $D^{\ell_1}$ and $D^{\ell_2}$ of \sothree\ on $\mathbb{C}^{2\ell_1+1}$ and $\mathbb{C}^{2\ell_2+1}$ respectively, the tensor-product group representation $D^{\ell_1} \otimes D^{\ell_2}$ of \sothree\ on $\mathbb{C}^{2\ell_1+1} \otimes \mathbb{C}^{2\ell_2+1}$ is defined such that $(D^{\ell_1} \otimes D^{\ell_2})(\rho)=D^{\ell_1}(\rho)\otimes D^{\ell_2}(\rho)$ for all $\rho \in \sothree$. Decomposing $D^{\ell_1} \otimes D^{\ell_2}$ into a direct sum of irreducible group representations then constitutes finding a change of basis for $\mathbb{C}^{2\ell_1+1} \otimes \mathbb{C}^{2\ell_2+1}$ such that $(D^{\ell_1} \otimes D^{\ell_2})(\rho)$ is block diagonal, where for each $\ell$ there is a block equal to $D^{\ell}(\rho)$. The necessary change of basis for $\hat{u}^{\ell_1} \otimes \hat{v}^{\ell_2} \in \mathbb{C}^{2\ell_1+1} \otimes \mathbb{C}^{2\ell_2+1}$ is given by
\begin{equation}\label{eqn:cgt-frag}
  (\hat{u}^{\ell_1} \otimes \hat{v}^{\ell_2})^{\ell}_m= \sum_{m_1=-\ell_1}^{\ell_1}\sum_{m_2=-\ell_2}^{\ell_2}C^{\ell_1, \ell_2, \ell}_{m_1,m_2,m}\hat{u}^{\ell_1}_{m_1}\hat{v}^{\ell_2}_{m_2},
\end{equation}
where $C^{\ell_1, \ell_2, \ell}_{m_1,m_2,m} \in \mathbb{C}$ denote Clebsch-Gordan coefficients whose symmetry properties are such that \mbox{$(\hat{u}^{\ell_1} \otimes \hat{v}^{\ell_2})^{\ell}_m$} is non-zero only for $|\ell_1-\ell_2| \leq \ell \leq \ell_1+\ell_2$. The use of \Eqref{eqn:cgt-frag} arises naturally in quantum mechanics when coupling angular momenta.

This property is useful since if $\hat{f}^{\ell_1} \in \mathbb{C}^{2\ell_1+1}$ and $\hat{f}^{\ell_2}\in \mathbb{C}^{2\ell_2+1}$ are two fragments that are equivariant with respect to rotations of the network input, then a rotation of $\rho$ applied to the network input results in $\hat{f}^{\ell_1} \otimes \hat{f}^{\ell_2}$ transforming as
\begin{align}
  [\hat{f}^{\ell_1} \otimes \hat{f}^{\ell_2}]^{\ell} & \mapsto [(D^{\ell_1}(\rho)\hat{f}^{\ell_1}) \otimes (D^{\ell_2}(\rho)\hat{f}^{\ell_2})]^{\ell} \\
                                                     & = [(D^{\ell_1} \otimes D^{\ell_2})(\rho)(\hat{f}^{\ell_1}\otimes \hat{f}^{\ell_2})]^{\ell}     \\ &= D^{\ell}(\rho)[\hat{f}^{\ell_1}\otimes \hat{f}^{\ell_2}]^{\ell},
\end{align}
where the final equality follows by block diagonality with respect to the chosen basis.  Therefore, if fragments $\hat{f}^{\ell_1}$ and $\hat{f}^{\ell_2}$ are equivariant with repsect to rotations of the network input, then so is the fragment $(C^{\ell_1, \ell_2, \ell})^{\top}(\hat{f}^{\ell_1} \otimes \hat{f}^{\ell_2}) \in \mathbb{C}^{2\ell+1}$, where we have written \Eqref{eqn:cgt-frag} more compactly. We now describe how \citet{kondor2018clebsch} use this fact to define equivariant non-linear transformations of elements in \fset.

A signal $f=\{\hat{f}_t^{\ell} \in \mathbb{C}^{2\ell+1}: \ell=0,..,L-1;\:  t=1,...,\tau_f^{\ell}\} \in \fset$ may be equivariantly and non-linearly transformed by an operator $\ntp: \fset \rightarrow \fset$ defined as
\begin{equation}\label{eqn:cgt}
  \ntp(f)=\{(C^{\ell_1, \ell_2, \ell})^{\top}(\hat{f}^{\ell_1}_{t_1} \otimes \hat{f}^{\ell_2}_{t_2})\: : \: \ell=0,...,L-1; \:(\ell_1, \ell_2)\in \plset; \:t_1=0,...,\tau_{f}^{\ell_1};\: t_2=0,...,\tau_{f}^{\ell_2}\},
\end{equation}
where for each degree $\ell \in \{0,...,L-1\}$ the set
\begin{equation}\label{eqn:plset-eqn}
  \plset = \{(\ell_1, \ell_2) \in \{0,...,L-1\}^2 : |\ell_1-\ell_2| \leq \ell \leq \ell_1+\ell_2 \}
\end{equation}
is defined in order to avoid the computation of trivially equivariant all-zero fragments.  We make the dependence on \plset\ explicit since we redefine it in \Secref{sec:efficient_gs2cnn}.

Unlike the pointwise activations discussed in the previous section this operator is strictly equivariant, with a mean relative equivariance error of $5.0 \times 10^{-7}$ (see Appendix~\ref{appendix:equivariance}). Note that applying this operator to signals on the sphere or rotation group results in generalized signals that are no longer on the sphere or rotation group. This is the rationale for the generalization to \fset: to unlock the ability to introduce non-linearity in a strictly equivariant manner.
Note, however, that $g=\ntp(f)$ has type $\tau_g=(\tau^0_g,...,\tau^{L-1}_g)$ where $\tau^{\ell}_{g}=\sum_{(\ell_1,\ell_2)\in\plset}\tau_{f}^{\ell_1}\tau_{f}^{\ell_2}$ and therefore application of this non-linear operator results in a drastic expansion in representation size, which is problematic.

\subsection{Generalized Spherical CNNs}

Equipped with operators to both linearly and non-linearly transform elements of \fset, with the latter also performing degree mixing, we may consider a network with representation spaces $\boldsymbol{\mathcal{H}}^{(0)}=...=\boldsymbol{\mathcal{H}}^{(S)}=\fset$.
We consider the $s$-th layer of the network to take the form of a triple $\mathcal{A}^{(s)} = (\mathcal{L}_1, \mathcal{N}, \mathcal{L}_2)$ such that
$\mathcal{A}^{(s)}(f^{(s-1)}) = \mathcal{L}_2(\mathcal{N}(\mathcal{L}_1(f^{(s-1)})))$,
where $\mathcal{L}_1, \mathcal{L}_2: \fset \rightarrow \fset$ are linear operators and $\mathcal{N}: \fset \rightarrow \fset$ is a non-linear activation operator.

The approaches of \citet{cohen2018spherical} and \citet{esteves2018learning} are encompassed in this framework as $\mathcal{A}^{(s)} = (\mathcal{L}^{(\psi)}, \mathcal{N}_{\sigma}, \mathcal{I})$, where $\mathcal{I}$ denotes the identity operator and filters $\psi$ may be defined to encode real-space properties such as localization (see Appendix~\ref{appendix:filters}). The framework of \cite{kondor2018clebsch} is also encompassed as $\mathcal{A}^{(s)} = (\mathcal{I},\mathcal{N}_{\otimes} ,\mathcal{L}^{(\psi)}_\text{G}) $. Here the generalized convolution $\mathcal{L}^{(\psi)}_\text{G}$ comes last to counteract the representation-expanding effect of the tensor-product activation and prevent it from compounding as signals pass through the network. Appendix~\ref{appendix:pws} lends intuition regarding relationships that may be captured by tensor-product activations followed by generalized convolutions.

For any intermediary representation $f^{(i)}\in\fset$ we may transition from equivariance with respect to the network input to invariance by discarding all but the scalar-valued fragments corresponding to $\ell=0$ (equivalent to average pooling for signals on the sphere and rotation group).
Finally, note that within this general framework we are free to consider hybrid approaches whereby layers proposed by \cite{cohen2018spherical,esteves2018learning,kondor2018clebsch} and others, and those presented subsequently, can be leveraged alongside each other within a single model.

%%% Local Variables:
%%% mode: latex
%%% TeX-master: "iclr2021"
%%% End:

\section{Efficient Generalized Spherical CNNs}\label{sec:efficient_gs2cnn}

Existing approaches to spherical convolutional layers that are encompassed within the above framework are computationally demanding.  They require the evaluation of costly harmonic transforms on the sphere and rotation group.  Furthermore, the only strictly rotationally equivariant non-linear layer is that of \cite{kondor2018clebsch}, which has an even greater computational cost, scaling with the fifth power of bandlimit  --- thereby limiting spatial resolution --- and quadratically with the number of fragments per degree --- thereby limiting representational capacity. This often prohibits the use of strictly equivariant spherical networks.

In this section we introduce a channel-wise structure, constrained generalized convolutions, and an optimized degree mixing set in order to construct new strictly equivariant layers that exhibit much improved scaling properties and parameter efficiency.  Furthermore, we adopt efficient sampling theory on the sphere and rotation group to achieve additional computational savings.

\subsection{Efficient Generalized Spherical Layers}

For an activation $f \in \fset$ the value $\bar{\tau}_f=\frac{1}{L}\sum_{\ell=1}^{L-1}\tau_f^{\ell}$ represents a resolution-independent proxy for its representational capacity. \cite{kondor2018clebsch} consider the separate fragments contained within $f$ to subsume the traditional role of separate channels and therefore control the capacity of intermediary network representations through specification of $\tau_f$. This is problematic because, whereas activation functions usually act on each channel separately and therefore have a cost that scales linearly with representational capacity (usually controlled by the number of channels), for the activation function \ntp\ not only does the cost scale quadratically with representational capacity $\bar{\tau}_f$, but so too does the size of $\ntp(f)$. This feeds forward the quadratic dependence to the cost of, and number of parameters required by, the proceeding generalized convolution.

More specifically, note that computation of $g=\ntp(f)$ requires the computation of $\sum_{\ell=0}^{L-1}\tau^{\ell}_g$ fragments, where \mbox{$\tau^{\ell}_{g}=\sum_{(\ell_1,\ell_2)\in\plset}\tau_{f}^{\ell_1}\tau_{f}^{\ell_2}$}. The size of $\plset$ is $\mathcal{O}(L\ell)$ for each $\ell$ and therefore the expanded representation has size $\sum_{\ell=0}^{L-1}\tau^{\ell}_g$, of order $\mathcal{O}(\bar{\tau}_f^2 L^3)$. By exploiting the sparsity of Clebsch-Gordan coefficients ($C^{\ell_1, \ell_2, \ell}_{m_1, m_2, m} = 0 \text{ if } m_1+m_2 \neq m$) each fragment $(C^{\ell_1, \ell_2, \ell})^{\top}(\hat{f}^{\ell_1} \otimes \hat{f}^{\ell_2})$ can be computed in $\mathcal{O}(\ell\min(\ell_1,\ell_2))$.  Hence, the total cost of computing all necessary fragments has complexity $\mathcal{O}(C^2 L^5)$, where $C=\bar{\tau}_f$ captures representational capacity.

\subsubsection{Channel-Wise Tensor-Product Activations}\label{sec:channel-wise_tpa}

As is more standard for CNNs we maintain a separate channels axis, with network activations taking the form $(f_1,...,f_K) \in \fsetk$ where $f_i \in \fset$ all share the same type $\tau_f$. The non-linearity \ntp\ may then be applied to each channel separately at a cost that is reduced by $K$-times relative to its application on a single channel with the same total number of fragments. This saving arises since for each $\ell$ we need only compute $K \sum_{(\ell_1,\ell_2)\in\plset}\tau_{f}^{\ell_1}\tau_{f}^{\ell_2}$ fragments rather than  $\sum_{(\ell_1,\ell_2)\in\plset}(K\tau_{f}^{\ell_1})(K\tau_{f}^{\ell_2})$.

\begin{figure}[b]
  \vspace{-3mm}
  \begin{subfigure}{0.815\textwidth}
    \includegraphics[width=\linewidth, left]{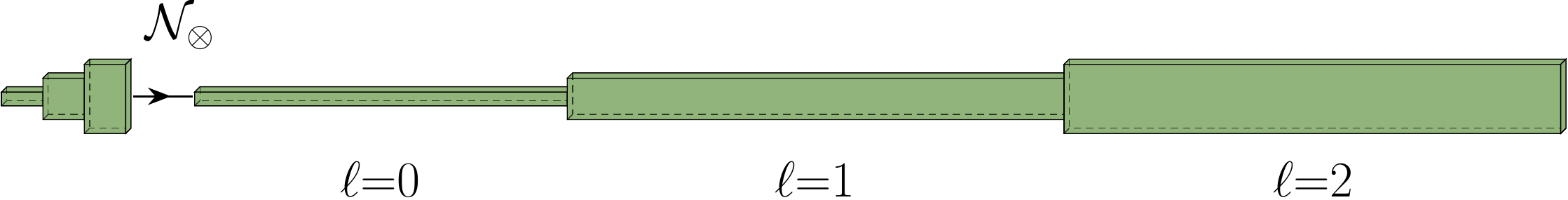}
    \caption{Prior approach to applying a tensor-product based non-linear operator}
  \end{subfigure}
  \begin{subfigure}{0.18\textwidth}
    \includegraphics[trim=-11 0 -11 0, width=\linewidth, right]{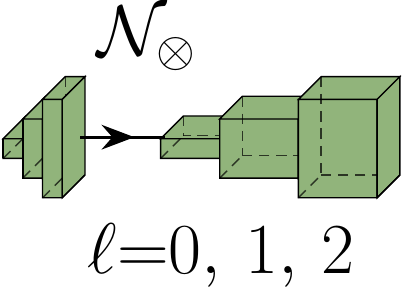}
    \caption{Ours}
  \end{subfigure}
  % \vspace{-1mm}
  \caption{Comparison (to scale) of the expansion caused by the tensor-product activation applied to inputs of equal representational capacity but different structure. With depth representing the number of channels and width the number of fragments for each degree, it is clear that by grouping fragments into $K$ separate channels the expansion (and therefore cost) can be $K$-times reduced. Visualization corresponds to inputs with $(L,K)$ equal to $(3,1)$ and $(3,3)$ for panel (a) and (b), respectively.}
  \label{fig:rep-diag}
  \vspace{-3mm}
\end{figure}

\mbox{Figure~\ref{fig:rep-diag}} visualizes this reduction for the case $K=3$. Note, however, that for practical applications $K \sim 100$ is more typical.  The $K$-times reduction in cost is therefore substantial and allows for intermediary activations with orders of magnitude more representational capacity.

By introducing this multi-channel approach and using $C=K$ rather than $C=\bar{\tau}_f$ to control representational capacity, we reduce the complexity of \ntp\ with respect to representational capacity from $\mathcal{O}(C^2)$ to $\mathcal{O}(C)$.

\subsubsection{Constrained Generalized Convolution}\label{sec:constrained_convs}

Although much reduced, for a signal $f \in \boldsymbol{\mathcal{F}}_L^{K_\text{in}}$ the channel-wise application of \ntp\ still results in a drastically expanded representation $g=\ntp(f)$, to which a representation-contracting generalized convolution must be applied in order to project onto a new activation $g' = \mathcal{L}_G^{(\psi)}(g)\in \boldsymbol{\mathcal{F}}_L^{K_\text{out}}$ of the desired type $\tau_{g'}$ and number of channels $K_\text{out}$. However, under our multi-channel structure computational and parameter efficiency can be improved significantly by decomposing $\mathcal{L}^{(\psi)}_\text{G}$ into three separate linear operators, $\mathcal{L}^{(\psi_1)}_{\text{G}_1}$, $\mathcal{L}^{(\psi_2)}_{\text{G}_2}$ and $\mathcal{L}^{(\psi_3)}_{\text{G}_3}$.

The first, $\mathcal{L}^{(\psi_1)}_{\text{G}_1}$, acts uniformly across channels, performing a linear projection down onto the desired type, and should be interpreted as a learned extension of \ntp\ which undoes the drastic expansion. The second, $\mathcal{L}^{(\psi_2)}_{\text{G}_2}$, then acts channel-wise, taking linear combinations of the (contracted number of) fragments within each channel. The third, $\mathcal{L}^{(\psi_3)}_{\text{G}_3}$, acts across channels, taking linear combinations to learn new features. More concretely, the three filters are of the form \mbox{$\psi_1 = \{\hat{\psi}_1^{\ell} \in \mathbb{C}^{\tau_{g}^{\ell} \times \tau_{g'}^{\ell}}: \ell=0,...,L-1\}$}, $\psi_2 = \{\hat{\psi}_2^{\ell,k} \in \mathbb{C}^{\tau_{g'}^{\ell} \times \tau_{g'}^{\ell}}: \ell=0,...,L-1; k=1,...,K_\text{in}\}$ and $\psi_3 = \{\hat{\psi}^{\ell}_3 \in \mathbb{C}^{K_\text{in} \times K_\text{out}}: \ell=0,...,L-1\}$, rather than a single filter of the form \mbox{$\psi = \{\hat{\psi}^{\ell} \in \mathbb{C}^{K_{\text{in}} \times \tau_{g}^{\ell} \times K_{\text{out}} \times \tau_{g'}^{\ell}}: \ell=0,...,L-1\}$}, leading to a large reduction in the number of parameters as $\tau_{g}^{\ell}$ is invariably very large.

By applying the first step uniformly across channels we minimize the parametric dependence on the expanded representation and allow new features to be subsequently learned much more efficiently. Together the second and third steps can be seen as analogous to depthwise separable convolutions often used in planar convolutional networks.
\subsubsection{Optimized Degree Mixing Sets}\label{sec:optimized_degree_mixing_sets}

We now consider approaches to reduce the $\mathcal{O}(L^5)$ complexity with respect to spatial resolution $L$.
In the definition of \ntp\ each element of \plset\ independently defines an equivariant fragment. Therefore a restricted \ntp\ in which only a subset of \plset\ is used for each degree $\ell$ still defines a strictly equivariant operator, while reducing computational complexity. In order to make savings whilst remaining at resolution $L$ it is necessary to consider subsets of \plset\ that scale better than $\mathcal{O}(L^2)$. The challenge is to find such subsets that do not hamper the ability of the resulting operator to inject non-linearity and mix information corresponding to different degrees $\ell$.

Whilst various subsetting approaches are possible, the following argument motivates an approach that we have found to be particularly effective.  If $(\ell_1,\ell_3) \in \plset$, then representational space is designated to capture the relationship between $\ell_1$ and $\ell_3$-degree information. However, if resources have been designated already to capture the relationship between $\ell_1$ and $\ell_2$-degrees, as well as between $\ell_2$ and $\ell_3$-degrees, then some notion of the relationship between $\ell_1$ and $\ell_3$-degrees has been captured already.  Consequently, it is unnecessary to designate further resources for this purpose.

More generally, consider the graph $G^{\ell}_L=(\mathbb{N}_L,\plset)$ with nodes $\mathbb{N}_L = \{0,...,L-1\}$ and edges \plset. A restricted tensor-product activation can be constructed by using a subset of \plset\ that corresponds to a subgraph of $G^{\ell}_L$.  The subgraph of $G^{\ell}_L$ captures some notion of the relationship between incoming $\ell_1$ and $\ell_2$-degree information if it contains a path between nodes $\ell_1$ and $\ell_2$. Therefore we are interested in subgraphs for which there exists a path between any two nodes if there exists such a path in the original graph, guaranteeing that any degree-mixing relationship captured by the original graph is also captured by the subgraph.

\begin{figure}[t]
  \vspace{-8mm}
  \begin{center}
    \begin{subfigure}{0.325\textwidth}
      \includegraphics[width=\linewidth]{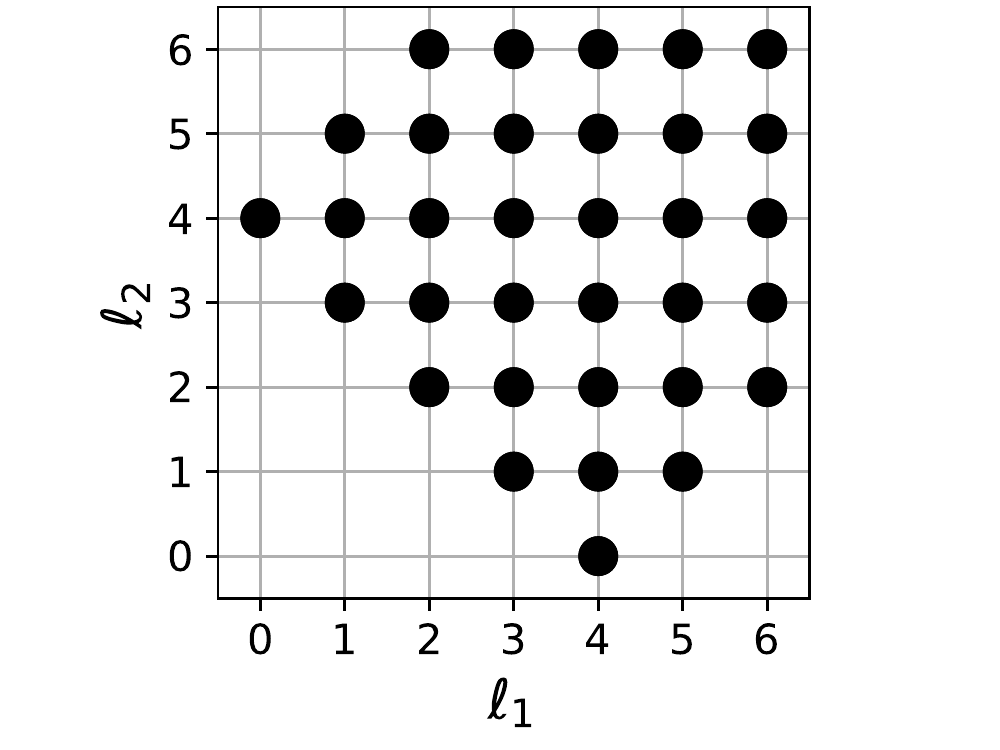}
      \vspace{-5mm}
      \caption{Full \plset\ set of size $\mathcal{O}(L^2)$}
    \end{subfigure}
    \begin{subfigure}{0.325\textwidth}
      \includegraphics[width=\linewidth]{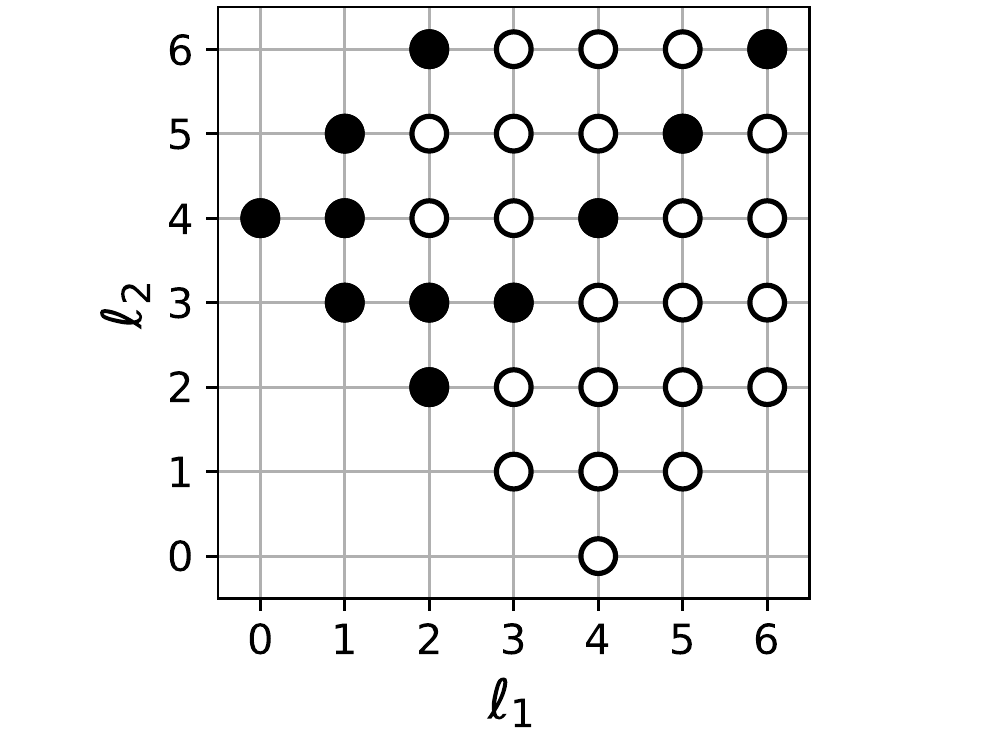}
      \vspace{-5mm}
      \caption{MST subset of size $\mathcal{O}(L)$}
    \end{subfigure}
    \begin{subfigure}{0.325\textwidth}
      \includegraphics[width=\linewidth]{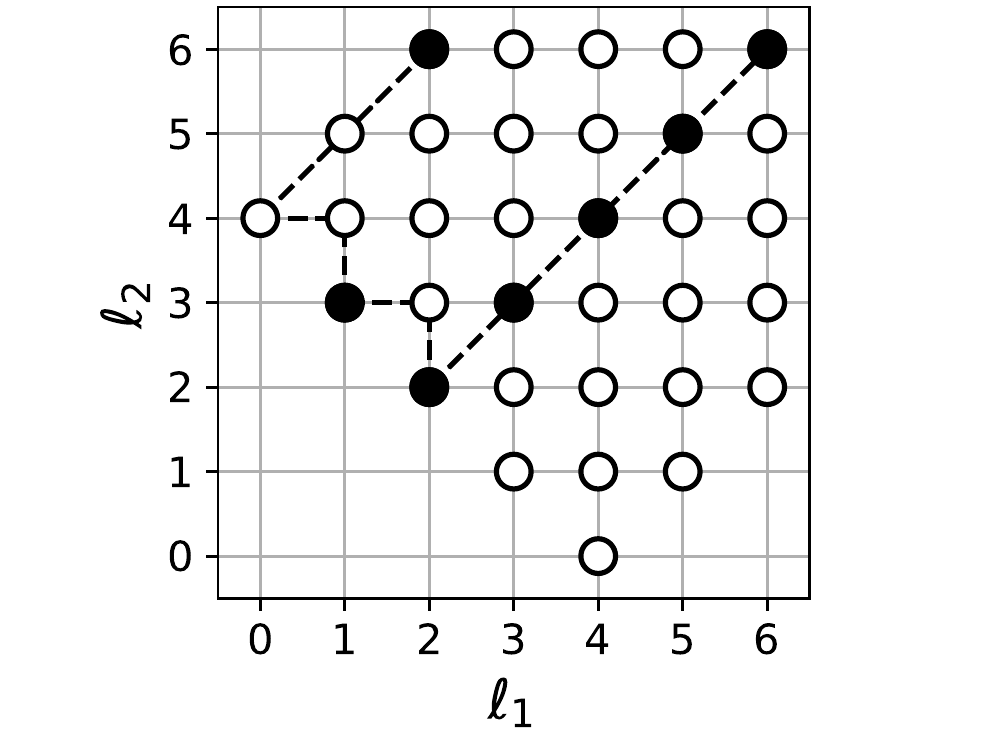}
      \vspace{-5mm}
      \caption{RMST subset of size $\mathcal{O}(\log L)$}\label{fig:pl-diags:rmst}
    \end{subfigure}
  \end{center}
  \vspace{-3mm}
  \caption{Visualization of the mixing set $\plset$  (for $L=7$ and $\ell=4$) and the approaches to subsetting based on the minimum spanning tree (MST) and reduced minimum spanning tree (RMST) mixing polices, which reduces related computation costs from $\mathcal{O}(L^2)$ to, respectively, $\mathcal{O}(L)$ or $\mathcal{O}(\log L)$.}
  \label{fig:pl-diags}
  \vspace{-2mm}
\end{figure}

The smallest subgraph satisfying this property is a minimum spanning tree (MST) of $G^{\ell}_L$. The set of edges corresponding to any MST has at most $L$ elements and we choose to consider its union with the set of loop-edges in $G^{\ell}_L$  (of the form $(\ell_1,\ell_1)$), which proved particularly important for injecting non-linearity. We denote the resulting set as $\bar{\mathbb{P}}^{\ell}_L$ and note that it satisfies $|\bar{\mathbb{P}}^{\ell}_L|\leq 2L$. Therefore the tensor-product activation $ \bar{\mathcal{N}}_{\otimes}$ corresponding to \Eqref{eqn:plset-eqn} with \plset\ replaced by $\bar{\mathbb{P}}^{\ell}_L$ has reduced spatial complexity $\mathcal{O}(L^4)$. Given that many minimal spanning trees of the unweighted graph $G^{\ell}_L$ exist for each $\ell$, we select the ones that minimize the cost of the resulting activation $\bar{\mathcal{N}}_{\otimes}$ by assigning to each edge $(\ell_1, \ell_2)$ in $G^{\ell}_L$ a weight equal to the cost of computing $(C^{\ell_1, \ell_2, \ell})^{\top}(\hat{f}^{\ell_1} \otimes \hat{f}^{\ell_2})$ and selecting the MST of the weighted graph.

An example of \plset\ and a MST subset $\bar{\mathbb{P}}^{\ell}_L$ is shown in Figure~\ref{fig:pl-diags}, where the dashed line in Figure~\ref{fig:pl-diags:rmst} shows the general form of the MST. Using this as a principled starting point we consider the further reduced MST (RMST) subset $\tilde{\mathbb{P}}^{\ell}_L$ corresponding to centering the MST at the edge $(\ell, \ell)$ and retaining only the edges that fall a distance of $2^{i}$ away on the dotted line for some $i \in \mathbb{N}$. We use $\tilde{\mathcal{N}}_{\otimes}$ to denote the corresponding operator and note that it has further reduced spatial complexity of $\mathcal{O}(L^3\log L)$.

We demonstrate in \Secref{sec:experiments} that networks that make use of the MST tensor-product activation achieve state-of-the-art performance.  Replacing the MST with RMST activation results in a small but insignificant degradation in performance, which is offset by the reduced computational cost.

\subsubsection{Reduction in Computational and Memory Footprints}
The three modifications proposed in Sections~\ref{sec:channel-wise_tpa} to \ref{sec:optimized_degree_mixing_sets} are motivated by improved scaling properties. Importantly, they also translate to large reductions in the computational and memory cost of strictly equivariant layers in practice, as detailed in Appendix~\ref{appendix:flops-mem}. Even at a modest bandlimit of $L=64$ and relatively small number of channels $K=4$, for example, the modifications together lead to a 51-times reduction in the number of flops required for computations and 16-times reduction in the amount of memory required to store representations, weights and gradients for training.

\subsection{Efficient Sampling Theory}\label{sec:efficient_sampling_theory}
By adopting sampling theorems on the sphere we provide access to underlying continuous signal representations that fully capture the symmetries and geometric properties of the sphere, and allow standard convolutions to be computed exactly and efficiently through their harmonic representations, as discussed in greater detail in Appendices~\ref{appendix:sphere} and \ref{appendix:convolutions}.
We adopt the efficient sampling theorems on sphere and rotation group of \citet{mcewen2011novel} and \citet{mcewen2015novel}, respectively, which reduce the Nyquist rate by a factor of two compared to those of \citet{driscoll:1994} and \citet{kostelec:2008}, which have been adopted in other spherical CNN constructions \citep[e.g.][]{cohen2018spherical, kondor2018clebsch, esteves2018learning, esteves2020spin}.  The sampling theorems adopted are equipped with fast algorithms to compute harmonic transforms, with complexity $\mathcal{O}(L^3)$ for the sphere and $\mathcal{O}(L^4)$ for the rotation group.  When imposing an azimuthal bandlimit $N \ll L$, the complexity of transforms on the rotation group can be reduced to $\mathcal{O}(N L^3)$, which we often exploit in our standard (non-generalized) convolutional layers.

%%% Local Variables:
%%% mode: latex
%%% TeX-master: "iclr2021"
%%% End:

\section{Experiments}\label{sec:experiments}

% Using our efficient generalized spherical CNN framework we construct networks that we apply to a number of spherical benchmark problems\blinded{ (implemented in our \code{\href{https://www.kagenova.com/products/fourpiAI/}{fourpiAI}}\footnote{\url{https://www.kagenova.com/products/fourpiAI/}} software package)}.  We achieve state-of-the-art performance, demonstrating
% the ability of our approach to enhance equivariance without compromising representational capacity or parameter efficiency.

Using our efficient generalized spherical CNN framework \blinded{ (implemented in the \code{\href{https://www.kagenova.com/products/fourpiAI/}{fourpiAI}}\footnote{\url{https://www.kagenova.com/products/fourpiAI/}} code)} we construct networks that we apply to numerous spherical benchmark problems.  We achieve state-of-the-art performance, demonstrating enhanced equivariance without compromising representational capacity or parameter efficiency.
In all experiments we use a similar architecture, consisting of 2--3 standard convolutional layers (e.g.\ \sphere\ or \sothree\ convolutions proceeded by ReLUs), followed by 2--3 of our efficient generalized layers. We adopt the efficient sampling theory described in \Secref{sec:efficient_sampling_theory} and encode localization of spatial filters as discussed in Appendix~\ref{appendix:filters}. Full  experimental details may be found in Appendix~\ref{appendix:experiments}.

\begin{table}[t]
    % \begin{footnotesize}
        \begin{minipage}{.55\linewidth}
    
                \begin{center}
                \normalsize
                \vspace{-3mm}
    
                \caption{Test accuracy for spherical MNIST digits classification problem}
                \footnotesize
                \vspace{-3mm}
    
                    \begin{tabular}{ l | c c c | c }
                        \label{table:mnist-results}
                        & NR/NR & R/R & NR/R & Params \\
                        \hline 
                        Planar CNN & 99.32 & 90.74 & 11.36 & 58k \\            
                        \cite{cohen2018spherical} & 95.59 & 94.62 & 93.40 & 58k \\
                        \cite{kondor2018clebsch} & 96.40 & 96.60 & 96.00 & 286k \\
                        \cite{esteves2020spin} & \textbf{99.37} & 99.37 & 99.08 & 58k \\
                        \hline
                        Ours (MST) & 99.35 & \textbf{99.38} & \textbf{99.34} & 58k \\
                        Ours (RMST) & 99.29 & 99.17 & 99.18 & \bf{57k} \\
                        \hline
                    \end{tabular}
                \end{center}
                \vspace{-3mm}
    
        \end{minipage}\hspace{.05\linewidth}%
        \begin{minipage}{.4\linewidth}
            \vspace{-.325cm}
                \begin{center}
                \normalsize
                \vspace{-3mm}
    
                \caption{Test root mean squared (RMS) error for QM7 regression problem}
                \footnotesize
                \vspace{-3mm}
    
                    \begin{tabular}{l|c|c}
                        \label{table:qm7-results}
                        & RMS & Params\\
                        \hline 
                        \cite{montavon2012atomenergyprediction} & 5.96 & - \\                        
                        \cite{cohen2018spherical} &  8.47 & 1.4M\\
                        \cite{kondor2018clebsch} &  7.97 & $>$1.1M\\
                        \hline
                        Ours (MST) & \bf{3.16} & 337k\\
                        Ours (RMST) & 3.46 & \bf{335k}\\
                        \hline
                    \end{tabular}
                \end{center}
                \vspace{-3mm}
    
        \end{minipage} 
    % \end{footnotesize}
    \end{table}

\begin{table}[t]
    \vspace{0mm}
    \caption{SHREC'17 object retrieval competition metrics (perturbed micro-all)}
    \label{table:shrec17-results}
    \begin{center}
        \vspace{-3mm}
        \footnotesize
        \begin{tabular}{l|ccccc|c}
            & P@N & R@N & F1@N & mAP & NDCG & Params\\
            \hline 
            \cite{kondor2018clebsch} & 0.707 & 0.722 & 0.701 & 0.683 & 0.756 & $>$1M \\
            \cite{cohen2018spherical} & 0.701 & 0.711 & 0.699 & 0.676 & 0.756 & 1.4M \\
            \cite{esteves2018learning} & 0.717 & \bf{0.737} & - & \bf{0.685} & - & 500k \\
            \hline
            Ours & \bf{0.719} & 0.710 & \bf{0.708} & 0.679 & \bf{0.758} & \bf{250k}\\
            \hline
        \end{tabular}
    \end{center}
    \vspace{-5mm}
\end{table}

\subsection{Rotated MNIST on the Sphere}

We consider the now standard benchmark problem of classifying MNIST digits projected onto the sphere. Three experimental modes NR/NR, R/R and NR/R are considered, indicating whether the training/test sets have been randomly rotated (R) or not (NR).  Results are presented in Table~\ref{table:mnist-results}, which shows that we closely match the prior state-of-the-art performance obtained by \citet{esteves2020spin} on the NR/NR and R/R modes, whilst outperforming all previous spherical CNNs on the NR/R mode, demonstrating the increased degree of equivariance achieved by our model.

Results are shown for models using both the MST-based and RMST-based mixing sets within the tensor-product activation. The results obtained when using the full sets \plset\ are very similar to those obtained when using the MST-based sets (e.g.\ full sets achieved an accuracy of $99.39$ for R/R).

\subsection{Atomization Energy Prediction}

We consider the problem of regressing the atomization energy of molecules given the molecule's Coulomb matrix and the positions of the atoms in space, using the QM7 dataset \citep{blum2009qm7, rupp2012qm7}.  Results are presented in Table~\ref{table:qm7-results}, which shows that we dramatically outperform other approaches, whilst using significantly fewer parameters.

\subsection{3D Shape Retrieval}

We consider the 3D shape retrieval problem on the SHREC'17 \citep{savva2017large} competition dataset, containing 51k 3D object meshes. We follow the pre-processing step of \cite{cohen2018spherical}, where several spherical projections of each mesh are computed, and use the official SHREC'17 data splits.  Results are presented in Table~\ref{table:shrec17-results} for the standard SHREC precision and recall metrics, which shows that we achieve state-of-the-art performance compared to other spherical CNN approaches, achieving the highest three of five performance metrics, whilst using significantly fewer parameters.

%%% Local Variables:
%%% mode: latex
%%% TeX-master: "iclr2021"
%%% End:

\section{Conclusions}

We have presented a generalized framework for CNNs on the sphere that encompasses various existing approaches.
We developed new efficient layers to be used as primary building blocks in this framework by introducing a channel-wise structure, constrained generalized convolutions, and optimized degree mixing sets determined by minimum spanning trees.
These new efficient layers exhibit strict rotational equivariance, without compromising on representational capacity or parameter efficiency.  When combined with the flexibility of the generalized framework to leverage the strengths of alternative layers, powerful hybrid models can be constructed.  On all spherical benchmark problems considered we achieve state-of-the-art performance, both in terms of accuracy and parameter efficiency.
In future work we intend to improve the scalability of our generalized framework further still.  In particular, we plan to introduce additional highly scalable layers, for example by extending scattering transforms \citep{mallat2012group} to the sphere, to further realize the potential of deep learning on a host of new applications where spherical data are prevalent.

%%% Local Variables:
%%% mode: latex
%%% TeX-master: "iclr2021"
%%% End:


\bibliography{efficient_generalized_s2cnn}

\begin{thebibliography}{32}
\providecommand{\natexlab}[1]{#1}
\providecommand{\url}[1]{\texttt{#1}}
\expandafter\ifx\csname urlstyle\endcsname\relax
  \providecommand{\doi}[1]{doi: #1}\else
  \providecommand{\doi}{doi: \begingroup \urlstyle{rm}\Url}\fi

\bibitem[Blum \& Reymond(2009)Blum and Reymond]{blum2009qm7}
Lorenz Blum and Jean-Louis Reymond.
\newblock 970 million druglike small molecules for virtual screening in the
  chemical universe database {GDB-13}.
\newblock \emph{Journal of the American Chemical Society}, 131:\penalty0 8732,
  2009.

\bibitem[Boomsma \& Frellsen(2017)Boomsma and Frellsen]{NIPS2017_6935}
Wouter Boomsma and Jes Frellsen.
\newblock Spherical convolutions and their application in molecular modelling.
\newblock In I.~Guyon, U.~V. Luxburg, S.~Bengio, H.~Wallach, R.~Fergus,
  S.~Vishwanathan, and R.~Garnett (eds.), \emph{Advances in Neural Information
  Processing Systems 30}, pp.\  3433--3443. Curran Associates, Inc., 2017.

\bibitem[Cohen et~al.(2018)Cohen, Geiger, K{\"o}hler, and
  Welling]{cohen2018spherical}
Taco Cohen, Mario Geiger, Jonas K{\"o}hler, and Max Welling.
\newblock Spherical {CNNs}.
\newblock In \emph{International Conference on Learning Representations}, 2018.
\newblock URL \url{https://arxiv.org/abs/1801.10130}.

\bibitem[Cohen et~al.(2019)Cohen, Weiler, Kicanaoglu, and
  Welling]{cohen2019gauge}
Taco Cohen, Maurice Weiler, Berkay Kicanaoglu, and Max Welling.
\newblock Gauge equivariant convolutional networks and the icosahedral {CNN}.
\newblock \emph{arXiv preprint arXiv:1902.04615}, 2019.
\newblock URL \url{https://arxiv.org/abs/1902.04615}.

\bibitem[Driscoll \& Healy(1994)Driscoll and Healy]{driscoll:1994}
James Driscoll and Dennis Healy.
\newblock Computing {F}ourier transforms and convolutions on the sphere.
\newblock \emph{Advances in Applied Mathematics}, 15:\penalty0 202--250, 1994.

\bibitem[Esteves(2020)]{esteves2020theoretical}
Carlos Esteves.
\newblock Theoretical aspects of group equivariant neural networks.
\newblock \emph{arXiv preprint arXiv:2004.05154}, 2020.
\newblock URL \url{https://arxiv.org/abs/2004.05154}.

\bibitem[Esteves et~al.(2018)Esteves, Allen-Blanchette, Makadia, and
  Daniilidis]{esteves2018learning}
Carlos Esteves, Christine Allen-Blanchette, Ameesh Makadia, and Kostas
  Daniilidis.
\newblock Learning {SO(3)} equivariant representations with spherical {CNNs}.
\newblock In \emph{Proceedings of the European Conference on Computer Vision
  (ECCV)}, pp.\  52--68, 2018.
\newblock URL \url{https://arxiv.org/abs/1711.06721}.

\bibitem[Esteves et~al.(2020)Esteves, Makadia, and Daniilidis]{esteves2020spin}
Carlos Esteves, Ameesh Makadia, and Kostas Daniilidis.
\newblock Spin-weighted spherical {CNNs}.
\newblock \emph{arXiv preprint arXiv:2006.10731}, 2020.
\newblock URL \url{https://arxiv.org/abs/2006.10731}.

\bibitem[Gallier \& Quaintance(2019)Gallier and Quaintance]{gallier2019aspects}
Jean Gallier and Jocelyn Quaintance.
\newblock \emph{Aspects of Harmonic Analysis and Representation Theory}.
\newblock 2019.
\newblock URL \url{https://www.seas.upenn.edu/~jean/nc-harmonic.pdf}.

\bibitem[Healy et~al.(2003)Healy, Rockmore, Kostelec, and Moore]{healy:2003}
Dennis Healy, Daniel Rockmore, Peter Kostelec, and S.~Moore.
\newblock {FFT}s for the 2-sphere -- improvements and variations.
\newblock \emph{Journal of Fourier Analysis and Applications}, 9\penalty0
  (4):\penalty0 341--385, 2003.

\bibitem[Jiang et~al.(2019)Jiang, Huang, Kashinath, Marcus, Niessner,
  et~al.]{jiang2019spherical}
Chiyu Jiang, Jingwei Huang, Karthik Kashinath, Philip Marcus, Matthias
  Niessner, et~al.
\newblock Spherical {CNN}s on unstructured grids.
\newblock \emph{arXiv preprint arXiv:1901.02039}, 2019.
\newblock URL \url{https://arxiv.org/abs/1901.02039}.

\bibitem[Kennedy \& Sadeghi(2013)Kennedy and Sadeghi]{kennedy2013hilbert}
Rodney~A Kennedy and Parastoo Sadeghi.
\newblock \emph{Hilbert space methods in signal processing}.
\newblock Cambridge University Press, 2013.

\bibitem[Kingma \& Ba(2015)Kingma and Ba]{kingma2015adam}
Diederik~P Kingma and Jimmy~Lei Ba.
\newblock Adam: A method for stochastic gradient descent.
\newblock In \emph{ICLR: International Conference on Learning Representations},
  2015.
\newblock URL \url{https://arxiv.org/abs/1412.6980}.

\bibitem[Kondor \& Trivedi(2018)Kondor and Trivedi]{kondor2018generalization}
Risi Kondor and Shubhendu Trivedi.
\newblock On the generalization of equivariance and convolution in neural
  networks to the action of compact groups.
\newblock In \emph{International Conference on Machine Learning}, pp.\
  2747--2755, 2018.
\newblock URL \url{https://arxiv.org/abs/1802.03690}.

\bibitem[Kondor et~al.(2018)Kondor, Lin, and Trivedi]{kondor2018clebsch}
Risi Kondor, Zhen Lin, and Shubhendu Trivedi.
\newblock {Clebsch-Gordan} nets: a fully fourier space spherical convolutional
  neural network.
\newblock In \emph{Advances in Neural Information Processing Systems}, pp.\
  10117--10126, 2018.
\newblock URL \url{https://arxiv.org/abs/1806.09231}.

\bibitem[Kostelec \& Rockmore(2008)Kostelec and Rockmore]{kostelec:2008}
Peter Kostelec and Daniel Rockmore.
\newblock {FFTs} on the rotation group.
\newblock \emph{Journal of Fourier Analysis and Applications}, 14:\penalty0
  145--179, 2008.

\bibitem[Mallat(2012)]{mallat2012group}
St{\'e}phane Mallat.
\newblock Group invariant scattering.
\newblock \emph{Communications on Pure and Applied Mathematics}, 65\penalty0
  (10):\penalty0 1331--1398, 2012.
\newblock URL \url{https://arxiv.org/abs/1101.2286}.

\bibitem[{Marinucci} \& {Peccati}(2011){Marinucci} and
  {Peccati}]{marinucci:2011:book}
Domenico {Marinucci} and Giovanni {Peccati}.
\newblock \emph{Random Fields on the Sphere: Representation, Limit Theorem and
  Cosmological Applications}.
\newblock Cambridge University Press, 2011.

\bibitem[McEwen \& Wiaux(2011)McEwen and Wiaux]{mcewen2011novel}
Jason McEwen and Yves Wiaux.
\newblock A novel sampling theorem on the sphere.
\newblock \emph{IEEE Transactions on Signal Processing}, 59\penalty0
  (12):\penalty0 5876--5887, 2011.
\newblock URL \url{https://arxiv.org/abs/1110.6298}.

\bibitem[McEwen et~al.(2007)McEwen, Hobson, Mortlock, and
  Lasenby]{mcewen:2006:fcswt}
Jason McEwen, Michael~P. Hobson, Daniel~J. Mortlock, and Anthony~N. Lasenby.
\newblock Fast directional continuous spherical wavelet transform algorithms.
\newblock \emph{IEEE Trans.\ Sig.\ Proc.}, 55\penalty0 (2):\penalty0 520--529,
  2007.
\newblock URL \url{https://arxiv.org/abs/astro-ph/0506308}.

\bibitem[McEwen et~al.(2013)McEwen, Vandergheynst, and
  Wiaux]{mcewen:2013:waveletsxv}
Jason McEwen, Pierre Vandergheynst, and Yves Wiaux.
\newblock On the computation of directional scale-discretized wavelet
  transforms on the sphere.
\newblock In \emph{Wavelets and Sparsity XV, SPIE international symposium on
  optics and photonics, invited contribution}, volume 8858, 2013.
\newblock URL \url{https://arxiv.org/abs/1308.5706}.

\bibitem[McEwen et~al.(2015{\natexlab{a}})McEwen, B{\"u}ttner, Leistedt,
  Peiris, and Wiaux]{mcewen2015novel}
Jason McEwen, Martin B{\"u}ttner, Boris Leistedt, Hiranya~V Peiris, and Yves
  Wiaux.
\newblock A novel sampling theorem on the rotation group.
\newblock \emph{IEEE Signal Processing Letters}, 22\penalty0 (12):\penalty0
  2425--2429, 2015{\natexlab{a}}.
\newblock URL \url{https://arxiv.org/abs/1508.03101}.

\bibitem[McEwen et~al.(2015{\natexlab{b}})McEwen, Leistedt, B{\"u}ttner,
  Peiris, and Wiaux]{mcewen:s2let_spin}
Jason McEwen, Boris Leistedt, Martin B{\"u}ttner, Hiranya Peiris, and Yves
  Wiaux.
\newblock Directional spin wavelets on the sphere.
\newblock \emph{IEEE Trans.\ Sig.\ Proc., submitted}, 2015{\natexlab{b}}.
\newblock URL \url{https://arxiv.org/abs/1509.06749}.

\bibitem[McEwen et~al.(2018)McEwen, Durastanti, and
  Wiaux]{mcewen:s2let_localisation}
Jason McEwen, Claudio Durastanti, and Yves Wiaux.
\newblock Localisation of directional scale-discretised wavelets on the sphere.
\newblock \emph{Applied Comput. Harm. Anal.}, 44\penalty0 (1):\penalty0 59--88,
  2018.
\newblock URL \url{https://arxiv.org/abs/1509.06767}.

\bibitem[Montavon et~al.(2012)Montavon, Hansen, Fazli, Rupp, Biegler, Ziehe,
  Tkatchenko, Lilienfeld, and M\"{u}ller]{montavon2012atomenergyprediction}
Gr\'{e}goire Montavon, Katja Hansen, Siamac Fazli, Matthias Rupp, Franziska
  Biegler, Andreas Ziehe, Alexandre Tkatchenko, Anatole~V. Lilienfeld, and
  Klaus-Robert M\"{u}ller.
\newblock Learning invariant representations of molecules for atomization
  energy prediction.
\newblock In F.~Pereira, C.~J.~C. Burges, L.~Bottou, and K.~Q. Weinberger
  (eds.), \emph{Advances in Neural Information Processing Systems 25}, pp.\
  440--448. Curran Associates, Inc., 2012.

\bibitem[Perraudin et~al.(2019)Perraudin, Defferrard, Kacprzak, and
  Sgier]{perraudin2019deepsphere}
Nathana{\"e}l Perraudin, Micha{\"e}l Defferrard, Tomasz Kacprzak, and Raphael
  Sgier.
\newblock Deepsphere: Efficient spherical convolutional neural network with
  {HEALPix} sampling for cosmological applications.
\newblock \emph{Astronomy and Computing}, 27:\penalty0 130--146, 2019.
\newblock URL \url{https://arxiv.org/abs/1810.12186}.

\bibitem[Rupp et~al.(2012)Rupp, Tkatchenko, M\"uller, and von
  Lilienfeld]{rupp2012qm7}
Matthias Rupp, Alexandre Tkatchenko, Klaus-Robert M\"uller, and O.~Anatole von
  Lilienfeld.
\newblock Fast and accurate modeling of molecular atomization energies with
  machine learning.
\newblock \emph{Physical Review Letters}, 108:\penalty0 058301, 2012.
\newblock URL \url{https://arxiv.org/abs/1109.2618}.

\bibitem[Savva et~al.(2017)Savva, Yu, Su, Kanezaki, Furuya, Ohbuchi, Zhou, Yu,
  Bai, Bai, et~al.]{savva2017large}
Manolis Savva, Fisher Yu, Hao Su, Asako Kanezaki, Takahiko Furuya, Ryutarou
  Ohbuchi, Zhichao Zhou, Rui Yu, Song Bai, Xiang Bai, et~al.
\newblock Large-scale 3d shape retrieval from shapenet core55: Shrec'17 track.
\newblock In \emph{Proceedings of the Workshop on 3D Object Retrieval}, pp.\
  39--50. Eurographics Association, 2017.

\bibitem[{Tegmark}(1996)]{tegmark:1996}
Max {Tegmark}.
\newblock An {Icosahedron-Based} method for pixelizing the celestial sphere.
\newblock \emph{Astrophys.\ J.\ Lett.}, 470:\penalty0 L81, October 1996.
\newblock URL \url{https://arxiv.org/abs/astro-ph/9610094}.

\bibitem[Thomas et~al.(2018)Thomas, Smidt, Kearnes, Yang, Li, Kohlhoff, and
  Riley]{thomas2018tensor}
Nathaniel Thomas, Tess Smidt, Steven Kearnes, Lusann Yang, Li~Li, Kai Kohlhoff,
  and Patrick Riley.
\newblock Tensor field networks: Rotation-and translation-equivariant neural
  networks for 3d point clouds.
\newblock \emph{arXiv preprint arXiv:1802.08219}, 2018.
\newblock URL \url{https://arxiv.org/abs/1802.08219}.

\bibitem[Trapani \& Navaza(2006)Trapani and Navaza]{trapani:2006}
Stefano Trapani and Jorge Navaza.
\newblock {Calculation of spherical harmonics and Wigner {\it d}~functions by
  {FFT}. Applications to fast rotational matching in molecular replacement and
  implementation into {\it AMoRe}}.
\newblock \emph{Acta Crystallographica Section A}, 62\penalty0 (4):\penalty0
  262--269, 2006.

\bibitem[Wandelt \& G\'{o}rski(2001)Wandelt and G\'{o}rski]{wandelt:2001}
Benjamin Wandelt and Krzysztof G\'{o}rski.
\newblock Fast convolution on the sphere.
\newblock \emph{Phys.\ Rev.\ D.}, 63\penalty0 (12):\penalty0 123002, 2001.
\newblock URL \url{https://arxiv.org/abs/astro-ph/0008227}.

\end{thebibliography}
\bibliographystyle{iclr2021_conference}

\vfill
\pagebreak

\appendix
\section{Representations of Signals on the Sphere and Rotation Group}
\label{appendix:sphere}

To provide further context for the discussion presented in the introduction and to elucidate the properties of different sampling theory on the sphere and rotation group, we concisely review representations of signals on the sphere and rotation group.

\subsection{Discretization}

It is well-known that a completely regular point distribution on the sphere does in general not exist \citep[e.g.][]{tegmark:1996}.  Consequently, while a variety of spherical discretization schemes exists (e.g.\ icosahedron, HEALPix, graph, and other representations), it is not possible to discretize (i.e.\ to sample or pixelize) the sphere in a manner that is invariant to rotations, i.e.\ a discrete sampling of rotations of the samples on the sphere will in general not map onto the same set of sample positions.  This differs to the Euclidean setting and has important implications when constructing convolution operators on the sphere, which clearly are a critical component of CNNs.

Since convolution operators are in general built using a translation operator -- equivalently a rotation operator when on the sphere -- it is thus not possible to construct a convolution operator directly on a discretized representation of the sphere that captures all of the symmetries of the underlying spherical manifold.  While approximate discrete representations can be considered, and are nevertheless useful, such representations cannot capture all underlying spherical symmetries.

\subsection{Sampling Theory}

Alternative representations, however, can capture all underlying spherical symmetries.  Sampling theories on the sphere \citep[e.g.][]{driscoll:1994,mcewen2011novel} provide a mechanism to capture all information content of an underlying continuous function on the sphere from a finite set of samples (and similarly on the rotation group; \citealt{kostelec:2008,mcewen2015novel}).  A sampling theory on the sphere is equivalent to a cubature (i.e.\ quadrature) rule for the exact integration of a bandlimited functions on the sphere.  While optimal cubature on the sphere remains an open problem, the most efficient sampling theory on the sphere and rotation group is that developed by \citet{mcewen2011novel} and \citet{mcewen2015novel}, respectively.

On a compact manifold like the sphere (and rotation group), harmonic (i.e.\ Fourier) space is discrete.  Hence, a finite set of harmonic coefficients captures all information content of an underlying continuous bandlimited signal.  Since such a representation provides access to the underlying continuous signal, all symmetries and geometric properties of the sphere are captured perfectly.  Such representations have been employed extensively in the construction of wavelet transforms on the sphere, where the use of sampling theorems on the sphere and rotation group yield wavelet transforms of discretized continuous signals that are theoretically exact \citep[e.g.][]{mcewen:2013:waveletsxv,mcewen:s2let_spin,mcewen:s2let_localisation}. Harmonic signal representations have also been exploited in spherical CNNs to access all underlying spherical symmetries and develop equivariance network layers \citep{cohen2018spherical,kondor2018clebsch,esteves2018learning,esteves2020spin}.

\subsection{Exact and Efficient Computation}

Signals on the sphere $f \in L^2(\sphere)$ may be decomposed into their harmonic representations as
\begin{equation}
    f(\omega) = \sum_{\ell=0}^{\infty}\sum_{m=-\ell}^{\ell} f^{\ell}_m Y^{\ell}_m(\omega) ,
\end{equation}
where their spherical harmonic coefficients are given by
\begin{equation}\label{eqn:spherical_harmonic_coeff}
    f^{\ell}_m
    = \langle f, Y^{\ell}_m \rangle
    = \int_\sphere \text{d}\mu(\omega) f(\omega) \, Y^{\ell\ast}_m(\omega),
\end{equation}
for $\omega \in \sphere$.
Similarly, signals on the rotation group $g \in L^2(\sothree)$  may be decomposed into their harmonic representations as
\begin{equation}\label{eqn:fourier_exp_so3}
    g(\rho) = \sum_{\ell=0}^{\infty}\frac{2\ell+1}{8\pi^2}\sum_{m=-\ell}^{\ell}\sum_{n=-\ell}^{\ell}g_{mn}^{\ell}D_{mn}^{\ell*}(\rho)
\end{equation}
where their harmonic (Wigner) coefficients are given by
\begin{equation}\label{eqn:wigner_coeff}
    g^{\ell}_{mn}
    = \langle g, D^{\ell\ast}_{mn} \rangle
    = \int_\sothree \text{d}\mu(\rho) g(\rho) \, D^{\ell}_{mn}(\rho),
\end{equation}
for $\rho \in \sothree$. Note that we adopt the convention where the conjugate of the Wigner $D$-function is used in \Eqref{eqn:fourier_exp_so3} since this leads to a convenient harmonic representation when considering convolutions \citep[cf.][]{mcewen2015novel,mcewen:s2let_localisation}.

As mentioned above, sampling theory pertains to strategies to capture all of the information content of band limited signals from a finite set of samples.  Since the harmonic space of the sphere and rotation group is discrete, this is equivalent to an exact quadrature rule for the computation of harmonic coefficients by \Eqref{eqn:spherical_harmonic_coeff} and \Eqref{eqn:wigner_coeff} from sampled signals.

The canonical equiangular sampling theory on the sphere was that developed by \citet{driscoll:1994}, and subsequently extended to the rotation group by \citet{kostelec:2008}. More recently, novel sampling theorems on the sphere and rotation group were developed by \citet{mcewen2011novel} and \citet{mcewen2015novel}, respectively, that reduce the Nyquist rate by a factor of two. Previous CNN constructions on the sphere  \citep[e.g.][]{cohen2018spherical, kondor2018clebsch, esteves2018learning, esteves2020spin} have adopted the more well-known sampling theories of \citet{driscoll:1994} and \citet{kostelec:2008}.  In contrast, we adopt the more efficient sampling theories of \citet{mcewen2011novel} and \citet{mcewen2015novel} to provide additional efficiency savings, implemented in the open source \code{\href{http://www.spinsht.org/}{ssht}}\footnote{\url{http://www.spinsht.org/}} and \code{\href{http://www.sothree.org/}{so3}}\footnote{\url{http://www.sothree.org/}} software packages (we also make use of a \mbox{TensorFlow} implementation of these algorithms in our private \code{tensossht}\blinded{\footnote{Available on request from \url{https://www.kagenova.com/}.}} code).  Note also that the sampling schemes associated with the theory of \citet{mcewen2011novel} (and other minor variants implemented in \code{ssht}) align more closely with the one-to-two aspect ratio of common spherical data, such as 360${}^\circ$ photos and videos.

All of the sampling theories discussed are equipped with fast algorithms to compute harmonic transforms, with complexity $\mathcal{O}(L^3)$ for transforms on the sphere \citep{driscoll:1994,mcewen2011novel} and complexity $\mathcal{O}(L^4)$ for transforms on the rotation group \citep{kostelec:2008,mcewen2015novel}.  Note that algorithms that achieve slightly lower complexity have been developed \citep{driscoll:1994,healy:2003,kostelec:2008} but these are known to suffer stability issues \citep{healy:2003,kostelec:2008}. By imposing an azimuthally bandlimit $N$, where typically $N \ll L$, the complexity of transforms on the rotation group can be reduced to $\mathcal{O}(N L^3)$ \citep{mcewen2015novel}, which we exploit in our networks.

These fast algorithms to compute harmonic transforms on the sphere and rotation group can be leveraged to yield the exact and efficient computation of convolutions through their harmonic representations (see Appendix~\ref{appendix:convolutions}).  By computing convolutions in harmonic space, pixelization and quadrature errors are avoided and computational complexity is reduced to the cost of the respective harmonic transforms.

\section{Convolution on the Sphere and Rotation Group}\label{appendix:convolutions}

For completeness we make explicit the standard (non-generalized) convolution operations on the sphere and rotation group that we adopt.  The general form of convolution for signals $f \in L^2(\Omega)$ either on the sphere ($\Omega=\sphere$) or rotation group ($\Omega =\sothree$) is specified by \Eqref{eqn:g-conv}, with harmonic representation given by \Eqref{eqn:conv}.  Here we provide specific expressions for the convolution for a variety of cases, describe the normalization constants that arise and may be absorbed into learnable filters, and derive the corresponding harmonic forms.  In practice all convolutions are computed in harmonic space since the computation is then exact, avoiding pixelisation or quadrature errors, and efficient when fast algorithms to compute harmonic transforms are exploited (see Appendix~\ref{appendix:sphere}).

\subsection{Convolution on the Sphere}

Given two spherical signals $f, \psi \in L^2(\sphere)$ their convolution, which in general is a signal on the rotation group, may be decomposed as
\begin{align}
  ( f \star \psi )(\rho) & = \langle f, R_{\rho} \psi \rangle                                                                                  \\
                         & = \int_{S^2} d\Omega(\omega) f(\omega) \psi^*(\rho^{-1}\omega)                                                      \\
  % &= \int_{S^2} d\Omega(\omega) \left[ \sum_{\ell m} f^{\ell}_m Y^{\ell}_m(\omega) \right]
  %  \left[\sum_{\ell' m'} \sum_n D_{m'n}^{\ell'*}(\rho)\psi^{\ell'*}_n Y^{\ell' *}_{m'}(\omega) \right] \\
                         & = \sum_{\ell m} \sum_{\ell' m'} \sum_n  f^{\ell}_m D_{m'n}^{\ell'*}(\rho)\psi^{\ell'*}_n \int_{S^2} d\Omega(\omega)
  Y^{\ell}_m(\omega)Y^{\ell' *}_{m'}(\omega)                                                                                                   \\
                         & = \sum_{\ell m} \sum_{\ell' m'} \sum_n  f^{\ell}_m D_{m'n}^{\ell'*}(\rho)\psi^{\ell'*}_n
  \delta_{\ell \ell'}\delta_{mm'}                                                                                                              \\ \label{eqn:conv:crucial-line}
                         & = \sum_{\ell m n} \left( f^{\ell}_m \psi^{\ell*}_n \right)D_{mn}^{\ell*}(\rho),
\end{align}
yielding harmonic coefficients
\begin{equation}
  (f * \psi)^{\ell}_{mn}= \frac{8\pi^2}{2\ell+1}f^\ell_m \psi^{\ell *}_n.
\end{equation}
The constants ${8\pi^2}/{(2\ell+1)}$ may be absorbed into learnable parameters.

\subsection{Convolution on the Sphere with Axisymmetric Filters}\label{appendix:convolutions:s2_axi}

When convolving a spherical signal $f \in L^2(\sphere)$ with an axisymmetric spherical filter $\psi \in L^2(\sphere)$ that is invariant to azimuthal rotations, the resultant $(f\star\psi)$ may be interpreted as a signal on the sphere. To see this note that an axisymmetric filter $\psi$ has harmonic coefficients $\psi^{\ell}_n = \psi^{\ell}_0 \delta_{n0}$ that are non-zero only for $m=0$.
Denoting rotations by their $zyz$-Euler angles $\rho = (\alpha, \beta, \gamma)$ and substituting into \Eqref{eqn:conv:crucial-line} we see that the convolution may be decomposed as
\begin{align}
  ( f \star \psi )(\alpha, \beta, \gamma) & = \sum_{\ell m n} \left( f^{\ell}_m \psi^{\ell *}_0 \delta_{n0} \right)D_{mn}^{\ell*}(\alpha, \beta, \gamma) \\
                                          & = \sum_{\ell m} f^{\ell}_m \psi^{\ell *}_0 D_{m0}^{\ell*}(\alpha, \beta, 0)                                  \\
                                          & = \sum_{\ell m} f^{\ell}_m \psi^{\ell *}_0\sqrt{\frac{4\pi}{2\ell+1}}Y^{\ell}_m(\beta, \alpha).
\end{align}
We may therefore interpret $(f\star\psi)$ as a signal on the sphere with spherical harmonic coefficients
\begin{equation}
  ( f \star \psi)^{\ell}_m = \sqrt{\frac{4\pi}{2\ell+1}}f^{\ell}_m \psi^{\ell *}_0.
\end{equation}
The constants $\sqrt{{4\pi}/{(2\ell+1)}}$ may be absorbed into learnable parameters.

\subsection{Convolution on the Rotation Group}\label{appendix:convolutions:so3}

Given two signals $f, \psi \in L^2(\sothree)$ on the rotation group their convolution may then be decomposed as
\begin{align}
  ( f \star \psi )(\rho) & = \langle f, R_{\rho} \psi \rangle                                                                                                                                                                                                          \\
                         & = \int_{\sothree} d\mu(\rho') f(\rho')\psi^*(\rho^{-1} \rho')                                                                                                                                                                               \\
                         & = \int_{\sothree} d\mu(\rho') \bigg[ \sum_{\ell}\frac{2\ell+1}{8\pi^2}\sum_{mn} f_{mn}^{\ell}D_{mn}^{\ell*}(\rho') \bigg] \bigg[ \sum_{\ell'}\frac{2\ell'+1}{8\pi^2}\sum_{m'n'}\psi_{m'n'}^{\ell'*}D_{m'n'}^{\ell'}(\rho^{-1}  \rho) \bigg] \\
                         & = \sum_{\ell} \frac{2\ell+1}{8\pi^2} \sum_{mn} f_{mn}^{\ell}\sum_{\ell'}\frac{2\ell'+1}{8\pi^2}\sum_{m'n'}\psi_{m'n'}^{\ell'*}
  \int_{\sothree}d\mu(\rho')D_{mn}^{\ell*}(\rho')D_{m'n'}^{\ell'}(\rho^{-1}  \rho')                                                                                                                                                                                    \\
                         & = \sum_{\ell} \frac{2\ell+1}{8\pi^2} \sum_{mn} f_{mn}^{\ell}\sum_{\ell'}\frac{2\ell'+1}{8\pi^2}\sum_{m'n'}\psi_{m'n'}^{\ell'*}
  \int_{\sothree}d\mu(\rho') D_{mn}^{\ell*}(\rho') \sum_k D_{km'}^{\ell'*}(\rho)D_{kn'}^{\ell'}(\rho') \label{eqn:wignerdinv}                                                                                                                                          \\
  %   &= \sum_{\ell} \frac{2\ell+1}{8\pi^2} \sum_{mn} f_{mn}^{\ell}\sum_{\ell'}\frac{2\ell'+1}{8\pi^2}\sum_{m'n'}\psi_{m'n'}^{\ell'*} 
  %   \sum_k D_{km'}^{\ell'*}(\rho) \int_{\sothree}d\mu(\rho') D_{mn}^{\ell*}(\rho') D_{kn'}^{\ell'}(\rho') \\
                         & = \sum_{\ell} \frac{2\ell+1}{8\pi^2} \sum_{mn} f_{mn}^{\ell}\sum_{\ell'}\frac{2\ell'+1}{8\pi^2}\sum_{m'n'}\psi_{m'n'}^{\ell'*}
  \sum_k D_{km'}^{\ell'*}(\rho) \frac{8\pi^2}{2\ell+1} \delta_{\ell\ell'}\delta_{mk}\delta_{nn'}                                                                                                                                                                       \\
  %   &= \sum_{\ell} \frac{2\ell+1}{8\pi^2}\sum_{mn}f_{mn}^{\ell}\sum_{m'}\psi_{m'n}^{\ell*}D_{mm'}^{\ell*}(\rho) \\
                         & = \sum_{\ell mm'} \frac{2\ell+1}{8\pi^2}D_{mm'}^{\ell*}(\rho) \bigg( \sum_n f_{mn}^{\ell}\psi_{m'n}^{\ell*} \bigg),
\end{align}
where for \Eqref{eqn:wignerdinv} we make use of the relation \citep[e.g.][]{marinucci:2011:book,mcewen:s2let_localisation}
\begin{align}
  D^{\ell}_{mn}(\rho^{-1} \rho') = \sum_k D^{\ell *}_{km}(\rho)D^{\ell}_{kn}(\rho').
\end{align}
This decomposition yields harmonic coefficients
\begin{equation}
  (f * \psi)^{\ell}_{m n} = \sum_{m'}f^\ell_{mm'} \psi^{\ell *}_{nm'}.
\end{equation}

%%% Local Variables:
%%% mode: latex
%%% TeX-master: "../iclr2021"
%%% End:

\section{Filters on the Sphere and Rotation Group }\label{appendix:filters}

When defining filters we look to encode desirable real-space properties, such as locality and regularity. However, in practice considerable computation may be saved by defining the filters in harmonic space and saving the cost of harmonic transforming ahead of harmonic space convolutions. We describe here how filters motivated by their real space properties may be defined directly in harmonic space.

\subsection{Dirac Delta Filters on the Sphere}

Spherical filters may be constructed as a weighted sum of Dirac delta functions on the sphere. This construction is useful as the harmonic representation has an analytic form that may be computed efficiently.  Furthermore, various real space properties can be encoded through sensible placement of the Dirac delta functions.

The spherical Dirac delta function $\delta_{\omega'}$ centered at $\omega' = (\theta', \phi') \in \sphere$ is defined as
\begin{equation}
  \delta_{\omega'}(\omega) = \frac{1}{\sin \theta} \delta_{\mathbb{R}}(\cos\theta - \cos\theta') \delta_{\mathbb{R}}(\phi - \phi'),
\end{equation}
where $\delta_{\mathbb{R}}$ is the familiar Dirac delta function on the reals centered at $0$. The Dirac delta on the sphere may be represented in harmonic space by
\begin{equation}
  (\delta_{\omega'})^\ell_{m} = Y^{\ell*}_{m}(\omega') =N^{\ell}_m\aleg{m}{\ell}{\cos{\theta'}}e^{-im\phi'},
\end{equation}
which follows form the sifting property of the Dirac delta, and where $Y^{\ell}_m$ denote the spherical harmonic functions, $\aleg{m}{\ell}{x}$ are associated Legendre functions and
\begin{equation}
  N^{\ell}_m=\sqrt{\frac{2\ell+1}{4\pi}\frac{(l-m)!}{(l+m)!}}
\end{equation}
is a normalizing constant.

This representation may then be used to define a filter $\psi \in L^2(\sphere)$ as a weighted sum of spherical Dirac delta functions, with weights $w_{ij}$ assigned to Dirac delta functions centered at points $\{(\theta_i, \phi_j): i=1,...,N_{\theta}; \;j=1,...,N_{\phi}\}$. The associated harmonic space representation is given by
\begin{align}
  \psi^\ell_{m} & = \sum_{i,j} w_{ij} N^{\ell}_m\aleg{m}{\ell}{\cos{\theta_i}}e^{-im\phi_j} \label{eqn:s2_dirac} \\
                & = \sum_{i} N^{\ell}_m\aleg{m}{\ell}{\cos{\theta_i}} \sum_j w_{ij} e^{-im\phi_j},
  \label{eqn:s2_dirac_final}
\end{align}
where fast Fourier transforms may be leveraged to compute the inner sum if the Dirac deltas are spaced evenly azimuthally (e.g.\ if $\phi_j={2\pi j}/{N_{\phi}}$).  Alternative arbitrary samplings can of course be considered if useful for a problem at hand.

When defining filters in this manner one should be careful not to over-parametrize by assigning more weights than needed to define a filter at the harmonic bandlimit of the signal with which we wish to convolve. For example, if the filter is to be convolved with a signal bandlimited at $L$ then a maximum of $2L-1$ Dirac deltas should be placed along each ring of constant $\theta$. One may also choose to interpolate the weights from a smaller number of learnable parameters acting as anchor points, allowing higher resolution filters to be defined with fewer learnable parameters.

\subsection{Dirac Delta Filters on the Rotation Group}

Similarly a Dirac delta function $\delta_{\rho'}$ on the rotation group \sothree\ centered at position \mbox{$\rho'=(\alpha', \beta', \gamma') \in \sothree$} is defined as
\begin{equation}
  \delta_{\rho'}(\rho) = \frac{1}{\sin \beta} \delta_{\mathbb{R}}(\alpha - \alpha') \delta_{\mathbb{R}}(\cos\beta - \cos\beta') \delta_{\mathbb{R}}(\gamma - \gamma'),
\end{equation}
with harmonic form
\begin{equation}
  (\delta_{\rho'})^{\ell}_{mn} = D^{\ell}_{mn}(\rho') =e^{-im\alpha'}d^{\ell}_{mn}(\beta')e^{-in\gamma'},
\end{equation}
where $d_{mn}^{\ell}$ are Wigner (small) $d$-matrices.

The filter $\psi \in L^2(\sothree)$ corresponding to a weighted sum of Dirac deltas with weights $w_{ijk}$ assigned to Dirac delta functions centered at points $\{(\alpha_i, \beta_j, \gamma_k): i=1,...,N_{\alpha}; \;j=1,...,N_{\beta}; \;k=1,...,N_{\gamma}\}$ has harmonic form
\begin{align}
  \psi^\ell_{m n} & = \sum_{i, j, k } w_{ijk} \: e^{-im\alpha_j}d^{\ell}_{mn}(\beta_i)e^{-in\gamma_k}\label{eqn:so3_dirac} \\
                  & = \sum_{j} d^{\ell}_{mn}(\beta_i) \sum_ie^{-im\alpha_j}\sum_kw_{ijk} e^{-in\gamma_k},
  \label{eqn:so3_dirac_final}
\end{align}
where again fast Fourier transforms may be leveraged to compute the inner two sums assuming the Dirac deltas are spaced evenly in $\alpha$ and $\gamma$. The outer sums of \Eqref{eqn:s2_dirac_final} and \Eqref{eqn:so3_dirac_final} can also be computed by fast Fourier transforms by decomposing the Wigner $d$-matrices into their Fourier representation \citep[cf.][]{trapani:2006,mcewen2011novel}. One should again be careful not to over-parametrize.

\section{Equivariance Tests}\label{appendix:equivariance}

To test rotational equivariance of operators we consider $N_f=100$ random signals $\{f_i\}_{i=1}^{N_f}$ in $L^2(\Omega_1)$ with harmonic coefficients sampled from the standard normal distribution and $N_{\rho}=100$ random rotations $\{\rho_j\}_{j=1}^{N_{\rho}}$ sampled uniformly on \sothree.  In order to measure the extent to which an operator $\mathcal{A}: L^2(\Omega_1) \rightarrow L^2(\Omega_2)$ is equivariant we evaluate the mean relative error
\begin{equation}
    d(\mathcal{A}(\mathcal{R}_{\rho_j} f_i), \mathcal{R}_{\rho_j} (\mathcal{A}f_i)) = \frac{1}{N_f}\frac{1}{N_{\rho}}\sum_{i=1}^{N_f}\sum_{j=1}^{N_\rho} \frac{\|\mathcal{A}(\mathcal{R}_{\rho_j} f_i) - \mathcal{R}_{\rho_j} (\mathcal{A}f_i))\|}{\|\mathcal{A}(\mathcal{R}_{\rho_j} f_i)\|}
\end{equation}
resulting from pre-rotation of the signal, followed by application of $\mathcal{A}$, as opposed to post-rotation after application of $\mathcal{A}$, where the operator norm $\|\cdot\|$ is defined using the inner product $\langle \cdot, \cdot \rangle_{L^2(\Omega_2)}$. 

Table~\ref{table:equivariance} presents the mean relative equivariance errors computed.  We consider the three standard convolutions described in Appendix~\ref{appendix:convolutions} (with a random filter $\psi_i$ for each signal $f_i$, generated in the same manner as $f_i$), the pointwise ReLU activation described in \Secref{sec:gs2cnn:pw} for signals on the sphere ($\Omega_1 = \sphere$) and rotation group ($\Omega_1 = \sothree$), and the composition of tensor-product activation with a generalized convolution, described in Sections \ref{sec:gs2cnn:tp} and \ref{sec:gs2cnn:gconv}, respectively. We follow the tensor-product activation with a generalized convolution in order to project down onto the sphere to allow the same notion of error to be adopted as for the other operators. For consistency with the context in which we leverage these operators, all experiments are performed using single-precision arithmetic.

We see that all three standard notions of convolution and the composition of the tensor-product activation and generalized convolution are all strictly equivariant to floating point machine precision, with errors on the order of $10^{-7}$. The pointwise ReLU operator is not strictly equivariant, with a mean relative error of $0.37$ for signals on the rotation group and $0.34$ for signals on the sphere. These errors reduce when the signals are oversampled before application of the ReLU, indicating that the error is due to aliasing induced by the spreading of information to higher degrees not captured at the original bandlimit. For example, for the pointwise ReLU operator on the rotation group oversampling by factors of $2\times, 4\times$ and $8\times$ results in a reduction in the mean relative equivariance error from $0.37$ to $0.098, 0.032$ and $0.0096$, respectively.

\begin{table}[h!]
    \caption{Layer equivariance tests}
    \label{table:equivariance}
    \vspace{-3mm}
    \begin{center}
        \begin{tabular}{lc}
            \multicolumn{1}{c}{Layer}  &\multicolumn{1}{c}{Mean Relative Error} 
            \\\hline 
            \sphere\ to \sphere\ conv.       & $4.4 \times 10^{-7}$  \\
            \sphere\ to \sothree\ conv.    &$5.3 \times 10^{-7}$ \\
            \sothree\ to \sothree\ conv.    &$9.3 \times 10^{-7}$ \\
            Tensor-product activation $\rightarrow$ Generalized conv.           & $5.0 \times 10^{-7}$ \\ \hline
            \sphere\ ReLU             &$3.4 \times 10^{-1}$ \\
            \sphere\ ReLU ($2\times$ oversampling)             &$8.9 \times 10^{-2}$ \\
            \sphere\ ReLU ($4\times$ oversampling)          &$2.9 \times 10^{-2}$ \\
            \sphere\ ReLU ($8\times$ oversampling)          &$1.3 \times 10^{-2}$ \\ \hline
            \sothree\ ReLU   &$3.7 \times 10^{-1}$ \\
            \sothree\ ReLU ($2\times$ oversampling)     &$9.8 \times 10^{-2}$ \\
            \sothree\ ReLU ($4\times$ oversampling)   &$3.2 \times 10^{-2}$ \\
            \sothree\ ReLU ($8\times$ oversampling)    &$9.6 \times 10^{-3}$ \\
        \end{tabular}
    \end{center}
\end{table}

\section{Connection Between the Tensor Product Activation and Pointwise Squaring}\label{appendix:pws}

To provide some intuition on the manner in which the tensor-product based activation introduces non-linearity into representations we describe its relationship to pointwise squaring for signals on the sphere. Here we consider the operator $\mathcal{N}: L^2(\sphere) \rightarrow L^2(\sphere)$ satisfying $(\mathcal{N}f)(x)=f^2(x)$ for all $x \in \sphere$, which differs subtly to $\mathcal{N}_{\sigma}$ with $\sigma(x)=x^2$ (using notation from \Secref{sec:gs2cnn:pw}), which corresponds to obtaining a sample-based representation at a finite bandlimit $L$ and applying the squaring at the sample positions. For the special case $\sigma(x)=x^2$ we can directly compute the harmonic representation corresponding to the equivariance-preserving continuous limit $L \rightarrow \infty$.

Given a spherical signal $f \in L^2(\sphere)$ with generalized representation $f=\{\hat{f}_0^{\ell} \in \mathbb{C}^{2\ell+1}: \ell=0,..,L-1\}$, the generalized representation of the signal $f^2 \in L^2(\sphere)$ is given as
\begin{equation}\label{hsq}
    f^2=\{ \sum_{(\ell_1, \ell_2) \in \mathcal{P}_{\ell,L}} (G^{\ell_1, \ell_2, \ell})^{\top}(\hat{f}^{\ell_1}_0 \otimes \hat{f}^{\ell_2}_0) \; : 
    \; \ell=0,1,...,L-1 \},
  \end{equation}
  where $G^{\ell_1, \ell_2, \ell} \in \mathbb{C}^{(2\ell_1+1) \times (2\ell_2+1) \times (2\ell+1)}$ are Gaunt coefficients defined as
  \begin{equation} 
    G^{j_1 j_2 j_3}_{m_1 m_2 m_3}=\int_{\sphere}d\mu(\omega)Y^{j_1}_{m_1}(\omega)Y^{j_2}_{m_2}(\omega)Y^{j_3*}_{m_3}(\omega).
  \end{equation}
  Gaunt coefficients are related to the Clebsch-Gordan coefficients by
  \begin{equation}
    G^{j_1 j_2 j_3}_{m_1 m_2 m_3} =w^{j_1j_2j_3}C^{j_1 j_2 j_3}_{m_1 m_2 m_3},
  \end{equation}
  where $w^{j_1j_2j_3}=(-1)^{m_3}\sqrt{\frac{(2j_1+1)(2j_2+1)}{4\pi(2j_3+1)}}C^{j_1 j_2 j_3}_{0 \; 0 \;0}$.
  Therefore, the continuous squaring operation corresponds to passing $f$ through a tensor-product activation \ntp\ followed by a generalized convolution back down onto the sphere (single fragment per degree) with weight assigned to the $(\ell_1, \ell_2)\in\plset$ fragment in degree-$\ell$ given by $w^{l_1l_2l_3}$.

This demonstrates that activations that are learnable within our framework can have very simple real-space interpretations. Even when confining outputs to the sphere we found it to be beneficial to allow the down-projection to be learnable rather than enforcing the weights given above for pointwise squaring. Learned activations will remain quadratic, however, given that output fragments are linear combinations of products between input fragments. 

\section{Comparison of Computational and Memory Footprints}\label{appendix:flops-mem}

\begin{figure}[t]
    % \vspace{-2mm}
    \begin{center}
        \begin{subfigure}{0.45\textwidth}
            \includegraphics[width=\linewidth]{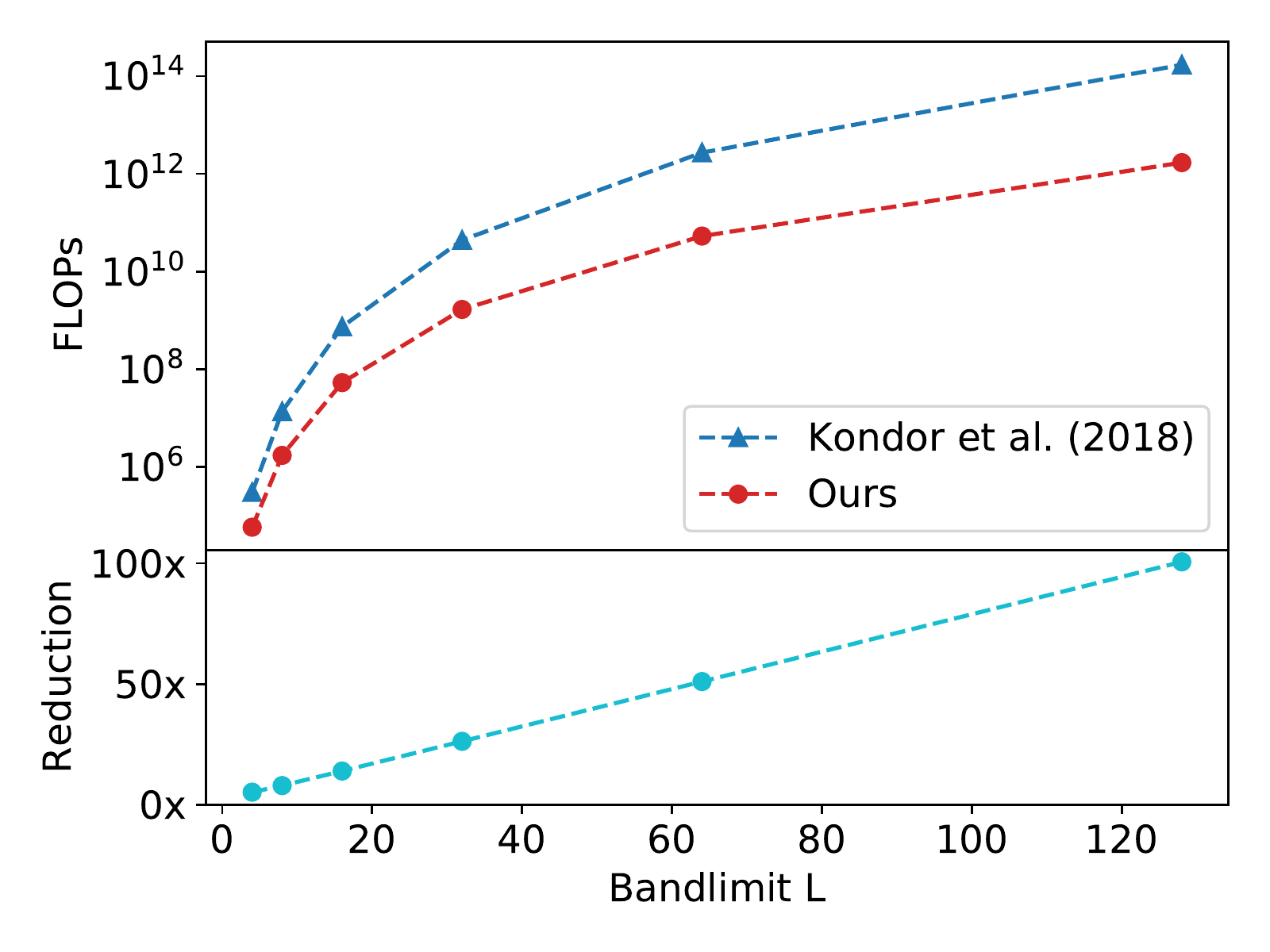}
            \vspace{-5mm}
            \caption{Computational cost}
        \end{subfigure}
        \hspace{1cm}
        \begin{subfigure}{0.45\textwidth}
            % \vspace{3mm}
            \includegraphics[width=\linewidth]{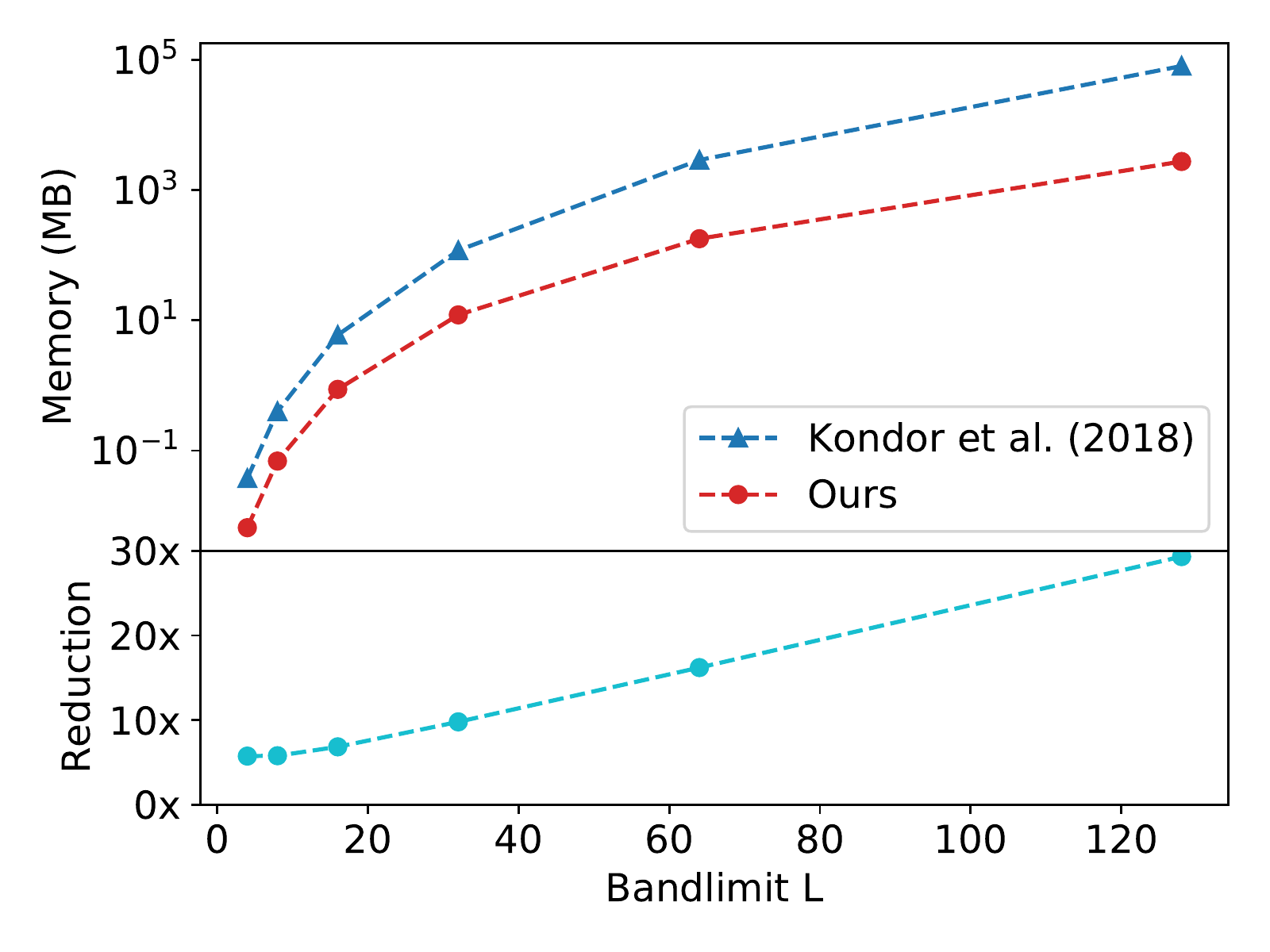}
            \vspace{-5mm}
            \caption{Memory requirements}
        \end{subfigure}
    \end{center}
    \vspace{-2mm}
    \caption{Comparison of computation and memory footprints between the generalized spherical CNN layers of \cite{kondor2018clebsch} and our efficient generalized layers. The reduction in cost due to our efficient layers is given as multiplicative factors in the lower plot of each panel.}
    \label{fig:flops-mem}
\end{figure}

We perform a quantitative analysis of the computational cost and memory requirements of our proposed layers, and comparisons to prior approaches, to demonstrate how the complexity savings made through our proposals in \Secref{sec:efficient_gs2cnn} translate to tangible efficiency improvements.

We consider the simplest comparison, between a multi-channel generalized signal $f=(f_1,...,f_K) \in \fsetk$ where each channel $f_i$ has type $\tau_{f_i}=(1,...,1)$ (and therefore corresponds to a signal on the sphere) and a uni-channel signal $g \in \fset$ of type $\tau^g=(K,...,K)$.  The setting corresponding to signal $f$ captures the efficiency improvements of our proposed layers, while the setting corresponding to $g$ represents the case without these improvements.  Notice that the total number of fragments is the same in both $f$ and $g$. We compare the number of floating point operations (FLOPs) and amount of memory required to perform a tensor-product based activation followed by a generalized convolution projecting back down onto a signal of the same type as the input. When applied to $f$, MST mixing sets $\bar{\mathbb{P}}^{\ell}_L$ (\Secref{sec:optimized_degree_mixing_sets}) and constrained generalized convolutions (\Secref{sec:constrained_convs}) are used. When applied to $g$ full mixing sets \plset and unconstrained generalized convolutions are used, as in \cite{kondor2018clebsch}. Considered are the costs for a single training instance (batch size of 1).

\Figref{fig:flops-mem} shows the computational costs and memory requirements, in terms of floating point operations and megabytes respectively, for $K=4$ and various spatial bandlimits $L$. We adopt the convention whereby complex addition and multiplication require 2 and 6 floating point operations respectively. At low bandlimits we see the saving arising from the channel-wise structure.  Note that the saving illustrated here is relatively small since $K=4$ for these experiments (so that the $L$ scaling is apparent), whereas in practice typically $K \sim 100$.  The saving then increases linearly in response to increases in the bandlimit of the input, as expected given the $\mathcal{O}(L^5)$ and $\mathcal{O}(L^4)$ spatial complexities. We see that even in this simple case, with a relatively small number of channels ($K=4$), both the computational and memory footprints are reduced by orders of magnitude. At a bandlimit of $L=128$ the computational cost is 101-times reduced and the memory requirement is 29-times reduced.

\section{Additional Information on Experiments}\label{appendix:experiments}

\subsection{Rotated MNIST on the Sphere}\label{appendix:experiments:mnist}

\begin{figure}[t!]
    \centering
    \vspace{-3mm}

    \includegraphics[trim=135 440 0 120, width=0.98\linewidth, left]{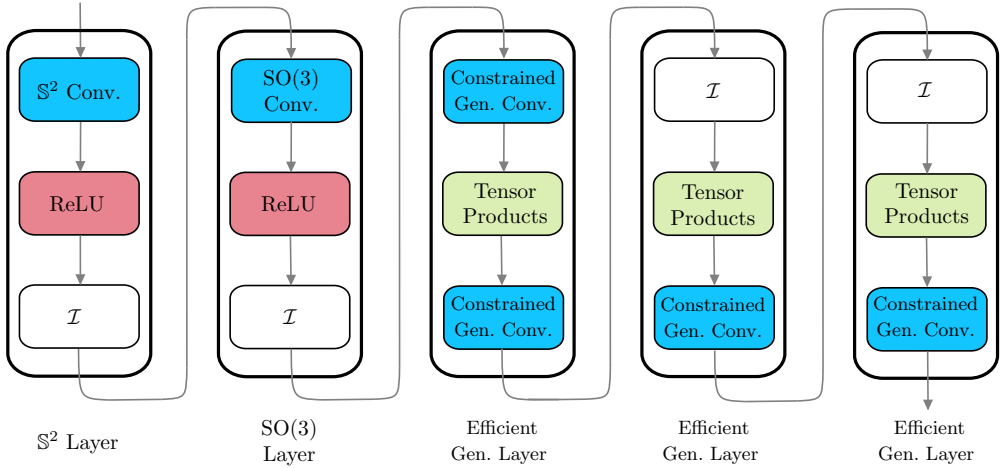}

    \vspace{-5mm}
    \caption{Visualization of the architecture used for the convolutional base in our hybrid models. The input to the first convolutional layer is a signal on the sphere. The output from the final convolutional layer are scalar values corresponding to fragments of degree $\ell=0$, which are then mapped through some fully connected layers to give the model output.}
    \vspace{-3mm}
    \label{fig:block_diagram}

\end{figure}

For our MNIST experiments we used a hybrid model with the architecture shown in Figure~\ref{fig:block_diagram}. The first block includes a directional convolution on the sphere that lifts the spherical input ($\tau_{f^{(0)}}^{\ell}=1$) onto the rotation group ($\tau_{f^{(1)}}^{\ell}=\min(2\ell+1, 2N_1-1)$).  The second block includes a convolution on the rotation group, hence its input and output both live on the rotation group. We then apply a restricted generalized convolution to map to type $\tau^{\ell}_{f^{(3)}}=\lceil {\tau_{\text{max}}}/{\sqrt{2\ell+1}} \rceil$, where $\tau_{\text{max}}{=}5$.  The same type is used for the following three channel-wise tensor-product activations and two restricted generalized convolutions until the final restricted generalized convolution maps down to a rotationally invariant representation ($\tau_{f^{(5)}}^{\ell}=\delta_{\ell0}$). As is traditional in convolution networks we gradually decrease the resolution, with $(L_0,L_1, L_2, L_3,L_4,L_5)=(20,10, 10, 6, 3, 1)$, and increase the number of channels, with $(K_0,K_1, K_2, K_3,K_4,K_5)=(1,20,22,24,26,28)$. We proceed these convolutional layers with a single dense layer of size $30$, sandwiched between two dropout layers (keep probability 0.5), and then fully connect to the output of size 10.

We train the network for $50$ epochs on batches of size $32$, using the Adam optimizer \citep{kingma2015adam} with a decaying learning rate starting at $0.001$. For the restricted generalized convolutions we follow the approach of \cite{kondor2018clebsch} by using $L_1$ regularization (regularization strength $10^{-5}$) and applying a restricted batch normalization across fragments, where the fragments are only scaled by their average and not translated (to preserve equivariance).

\subsection{Atomization Energy Prediction}\label{appendix:experiments:qm7}

When regressing the atomization energy of molecules there are two inputs to the model: the number of atoms of each element contained in the molecule; and spherical cross-sections of the potential energy around each atom. We adopt the high-level QM7-specific architecture of \citet{cohen2018spherical} which contains a spherical CNN as a sub-model, for which we substitute our own. This results in an overall model that is invariant to both rotations of the molecule around each constituent atom and to permutations of the ordering of the atoms. 

The first (non-spherical) input is mapped onto a scalar output using a multi-layer perceptron (MLP) with three hidden layers of sizes $100$, $100$ and $10$ (and ReLU activations). The second input, multiple spherical cross sections for each atom, are separately projected using a shared spherical CNN (of architecture described below) onto lower dimensional vectors of size $64$. The mean vector is then taken across atoms (ensuring invariance w.r.t. permutations of the atoms) and mapped onto a scalar output using an MLP with a single hidden layer of size $512$ (with a ReLU activation). The predicted energy is then taken to be the sum of the two scalar outputs.

As a starting point we train the first MLP to regress the atomization energies alone (achieving RMS ${\sim}20$), before pairing it with the spherical model (and its connected MLP). We then train the joint model for 60 epochs, again with the Adam optimizer, a decaying learning rate (starting at $2.5 \times 10^{-4}$), regularizing the efficient generalized layers with $L_2$ regularization (strength $2.5 \times 10^{-6}$) and batch sizes of $32$. 

For the spherical component we again adopt the convolutional architecture shown in Figure~\ref{fig:block_diagram} except with one fewer efficient generalized layer. We use bandlimits of $(L_0,L_1, L_2, L_3,L_4){=}(10, 6, 6, 3, 1)$, channels of $(K_0,K_1, K_2, K_3,K_4){=}(5,16,24,32,40)$ and $\tau_{\text{max}}=6$. One minor difference is that this time we include a skip connection between the $\ell=0$ components of the fourth and fifth layer.  We proceed the convolutional layers with two dense layers of size $(256, 64)$ and use batch normalization between each layer. 

\subsection{3D Shape Retrieval}\label{appendix:experiments:shrec17}

To project the 3D meshes of the SHREC'17 data onto spherical representations (bandlimited at $L=128$) we adopt the preprocessing approach of \cite{cohen2018spherical} and augment the data with random rotations and translations.

We construct a model with an architecture that is again similar to that described in Appendix~\ref{appendix:experiments:mnist} but with an additional axisymmetric convolutional layer prepended to the start of the network and one fewer efficient generalized layers. We use bandlimits $(L_{0}, L_{1},L_{2}, L_{3}, L_{4}, L_{5})=(128, 32, 16, 16, 6, 1)$, channels $(K_{0},K_{1},K_{2}, K_{3}, K_{4}, K_{5})=(6, 20, 30, 40, 60, 70)$ and $\tau_{\text{max}}=6$ for the efficient generalized layers. The convolutional layers are followed by a dense layer of size $128$ which is fully connected to the output (of size 55). 

We again train with the Adam optimizer, a decaying learning rate (starting at $5\times10^{-4}$) and batch sizes of $8$, this time until performance on the validation set showed no improvement for at least $4$ epochs ($36$ epochs in total). We perform batch normalization between convolutional layers and dropout preceding the dense layer. We regularize the efficient generalized layers with $L_2$ regularization (strength $10^{-5}$). When testing our model we average the output probabilities over 15 augmentations of the data.

\end{document}